\documentclass{article} 
\PassOptionsToPackage{table}{xcolor}
\usepackage{iclr2027_conference,times}

\usepackage{amsmath,amsfonts,bm}

\def\eqref#1{equation~\ref{#1}}

\def\1{\bm{1}}

\DeclareMathAlphabet{\mathsfit}{\encodingdefault}{\sfdefault}{m}{sl}
\SetMathAlphabet{\mathsfit}{bold}{\encodingdefault}{\sfdefault}{bx}{n}

\newcommand{\E}{\mathbb{E}}

\newcommand{\R}{\mathbb{R}}

\newcommand{\softmax}{\mathrm{softmax}}
\newcommand{\sigmoid}{\sigma}

\usepackage{hyperref}
\usepackage{url}
\usepackage{graphicx}
\usepackage{booktabs}
\usepackage{multirow}
\usepackage{adjustbox}
\definecolor{worldblue}{RGB}{25,115,195}
\definecolor{actionorange}{RGB}{220,115,45}
\newcommand{\model}{\textsc{Bridge-WA}}
\usepackage{booktabs}
\usepackage{multirow}
\usepackage{amssymb}
\providecommand{\method}{\textsc{Bridge-WA}}
\providecommand{\model}{\textsc{Bridge-WA}}
\providecommand{\wbblock}{WorldBridge}

\usepackage{fontawesome}
\usepackage{algpseudocode}
\usepackage{amsmath, amssymb}
\usepackage{newfloat}
\usepackage{listings}
\usepackage{hyperref}
\usepackage{color}
\usepackage{xspace}
\usepackage{booktabs}
\usepackage{makecell}
\usepackage{graphicx} 
\usepackage{bibentry}
\usepackage{xspace}
\usepackage{marvosym}
\usepackage{multirow}
\usepackage{caption}
\definecolor{academicpink}{RGB}{250,235,239}
\usepackage{multirow}
\usepackage{enumitem}
\usepackage{amsmath,amssymb,graphicx,booktabs,tabularx,longtable,mathtools}
\usepackage{threeparttable}
\usepackage{amsmath}
\usepackage{wrapfig}

\definecolor{sectionpink}{RGB}{253,240,240}
\definecolor{methodblue}{RGB}{238,247,251}
\definecolor{deltagreen}{RGB}{240,249,238}

\DeclareMathOperator{\Pool}{Pool}
\DeclareMathOperator{\LN}{LN}
\DeclareMathOperator{\sg}{sg}

\title{
  \texorpdfstring{
    \mbox{%
      \raisebox{-1em}{%
        \includegraphics[
          height=2.8em,
          trim=0 0 5 0,
          clip
        ]{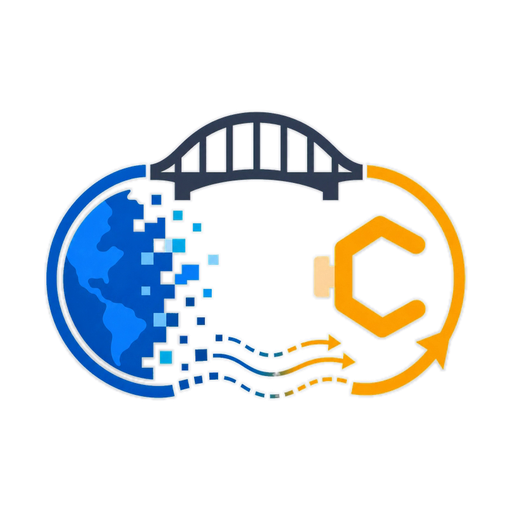}%
      }%
      \hspace{0.3em}%
      Bridge-WA:
    }
    Learning Action-Relevant World Dynamics for Robotic Manipulation
  }{
    Bridge-WA: Learning Action-Relevant World Dynamics for Robotic Manipulation
  }
}

\author{
Yongjie Bai$^{1,2}$
Mingtong Dai$^{2,3}$
Zhouxia Wang$^{1}$
Hanting Wang$^{1}$
Qijun Zhong$^{1}$
Feng Yan$^{5}$\\
~\textbf{Yang Liu$^{1,4,\textsuperscript{\Letter}}$ }
\textbf{Liang Lin$^{1,2,4}$}
\\
$^1$Sun Yat-sen University 
$^2$Pengcheng Laboratory \\
$^3$Shenzhen Institutes of Advanced Technology, Chinese Academy of Sciences
$^4$X-Era AI Lab \\
$^5$Xi'an Jiaotong University
}

\iclrfinalcopy

\begin{document}

\maketitle

\begingroup
\renewcommand{\thefootnote}{\Letter}
\footnotetext{
Corresponding Author
}
\endgroup

\fancyhead{}
\renewcommand{\headrulewidth}{0pt}

\begin{figure}[h]
    \centering
    \vspace{-18pt}
    \includegraphics[width=1\linewidth]{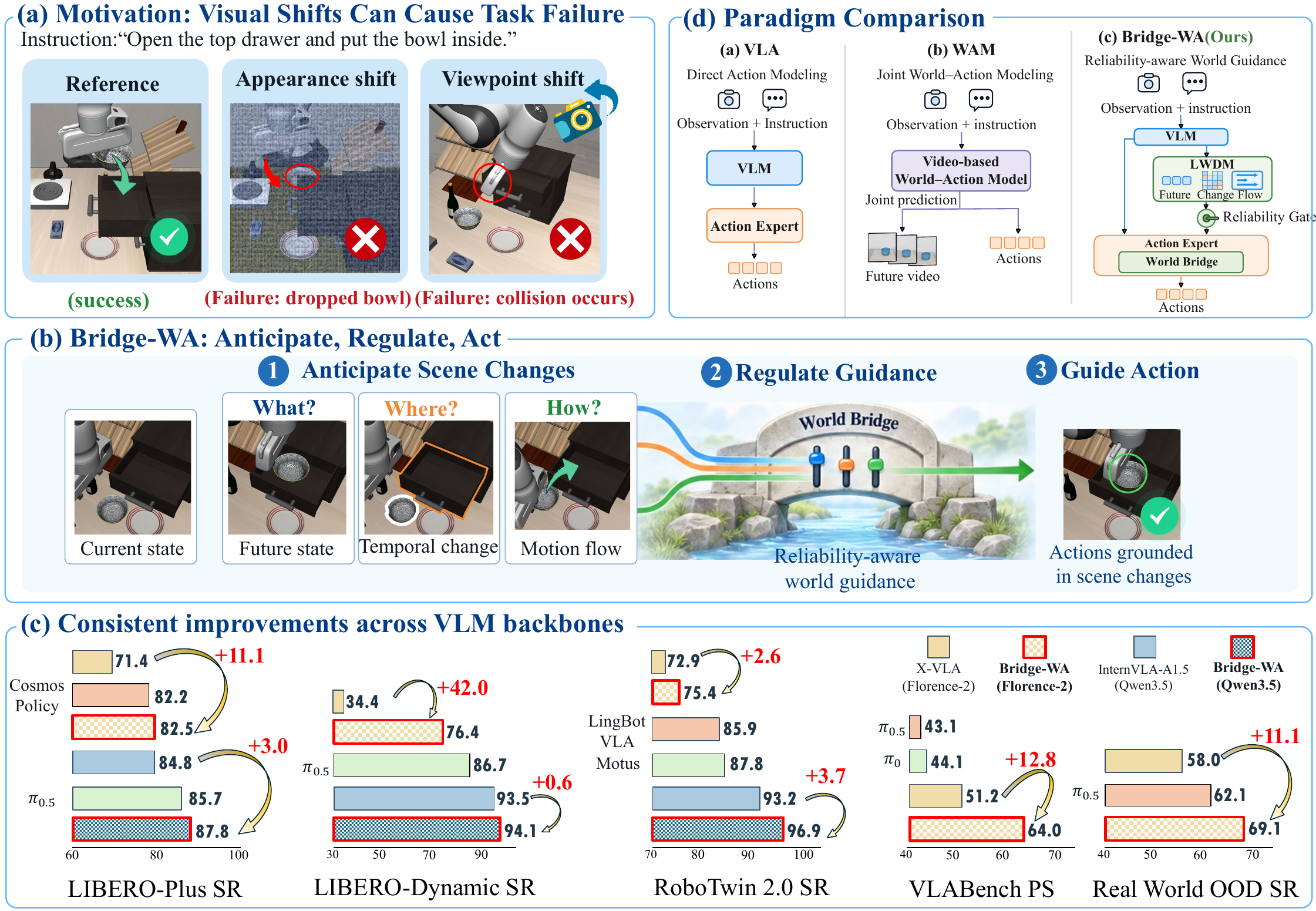}
    \vspace{-18pt}
    \caption{\textbf{(a)} Motivation: Appearance and viewpoint shifts can cause manipulation failures despite unchanged task semantics.
\textbf{(b)} In contrast, Bridge-WA anticipates \textbf{what}, \textbf{where}, and \textbf{how} the scene will change, regulates the reliability of these world priors, and grounds actions in the predicted changes, enabling successful task completion under visual shifts.
\textbf{(c)} \method{} consistently improves different VLM backbones across four simulation and real-world benchmarks.
\textbf{(d)} Unlike standard VLAs that directly predict actions and world-action models that jointly generate future video, \method{} injects reliability-aware world priors into the action expert.}
\vspace{8pt}
    \label{fig:banner}
\end{figure}

\begin{abstract}
General-purpose vision-language-action (VLA) models leverage large-scale vision-language priors to understand scenes and instructions, but primarily generate actions directly from the current observation. World-action models (WAMs) further model future scene states, offering a broader view of how the environment may evolve. However, for robotic manipulation, predicting the entire future scene can introduce information beyond what is necessary for action generation; more importantly, effective actions require knowing not only \textbf{what} the scene may become, but also \textbf{where} and \textbf{how} relevant changes unfold. To bridge action generation with these action-relevant aspects of the world, we present \textbf{\method{}}, a general world-action framework that learns complementary representations of future states, spatial changes, and local motion. Specifically, \method{} consists of a Latent World Dynamics Module (LWDM) and WorldBridge. LWDM predicts future states, spatial changes, and local motion from VLM outputs, supervised by corresponding world targets. The WorldBridge are embedded into the action transformer and inject layer-specific combinations of these world priors through multi-source attention, spatiotemporal biases, and reliability-gated feature modulation. This design grounds action generation in action-relevant future dynamics while adaptively regulating world guidance, enabling robust generalization to visual disturbances and viewpoint shifts. We evaluate \method{} on LIBERO-Plus, LIBERO-Dynamic, RoboTwin 2.0, VLABench, and real-world robots, where it achieves maximum success-rate improvements of 11.1\%, 42.0\%, 3.7\%, 23.4\%, and 11.1\%, respectively, and achieves state-of-the-art average success rates on LIBERO-Dynamic, RoboTwin 2.0 and real-world. In particular, \method{} demonstrates strong generalization to visual variations and viewpoint shifts in both simulation and real-world settings. Code and visualizations are available at: \faGithub\ \href{https://hcplab-sysu.github.io/BRIDGE-WA}{\method{}}.



\end{abstract}

\section{Introduction}
General-purpose manipulation requires more than recognizing objects and parsing instructions. A robot must also reason about intervention: after it pushes, grasps, pours, or inserts an object, which regions should change, how should they move, and which visual details should remain irrelevant? Recent embodied foundation models and generalist vision-language-action (VLA) models have advanced through large-scale, cross-embodiment robot training and the transfer of web-scale vision-language priors to robot control ~\citep{brohan2022rt1,oneill2024openx,octo2024,bjorck2025gr00t,zitkovich2023rt2,kim2024openvla,black2024pi0, li2024cogact,black2025pi05,team2025gemini, cai2026internvla1,ma2026internvla15,cheang2025gr3,bu2025univla,wang2026qwenvla,yuan2026qwenrobotmanip,zheng2025xvla,wu2026linbotvla}.
Action chunking and diffusion- or flow-based models further improve temporal consistency and continuous control~\citep{zhao2023act,chi2023diffusionpolicy,liu2025rdt,black2024pi0,kim2025fine,black2026real}. Yet most direct VLA models still learn a reactive mapping from current observation and language to actions, leaving future consequences weakly represented and making policies vulnerable to shortcut correlations in backgrounds, colors, and camera-specific appearance, thereby leading to action errors and task failures, as shown in the Fig~\ref{fig:banner}(a).

\vspace{-0.2em}
World models offer a natural way to close this gap: predict how the environment will evolve, then use that prediction to guide action. Latent dynamics, imagination-based agents, visual foresight, and robotic video prediction have shown that anticipated futures can support planning and manipulation~\citep{ha2018world,hafner2019planet,hafner2020dreamer,hafner2023dreamerv3,schrittwieser2020muzero,finn2017visualforesight,ebert2018visualforesight,babaeizadeh2021fitvid,du2023unipi,zhou2024robodreamer}. Recent world-action models extend this idea by jointly modeling future states and actions~\citep{hou2026worldmodelsurvey,wang2026worldactionmodels,luo2026beingh07,liu2026oawam,bi2026motus,yuan2026fastwam,kim2026cosmos,zhao2026faster1,ma2026faster2}. The challenge is practical: running or decoding a large generative world model can be expensive, and dense image/video prediction allocates capacity to many pixels that do not affect control. For manipulation, the useful future is often smaller than a photorealistic rollout: it is a structured \textbf{change field} that tells the policy where the intended interaction takes effect and how the affected regions move.

\vspace{-0.2em}


To address these challenges, We propose \method{}, a general world-action framework that conditions action generation on complementary world priors, as shown in the Fig~\ref{fig:banner}(b). During training, we construct supervision directly from observed futures in robot demonstrations. Frozen WAN VAE ~\citep{wan2025wan} and DINOv3~\citep{simeoni2025dinov3} encoders provide future tokens $T_f$ that capture complementary latent and visual structure, while differences between current and future features yield change maps $M_c$ that highlight regions of scene change. Image-space optical flow provides motion maps $M_o$ describing local displacement direction and magnitude. A Latent World Dynamics Module (LWDM) learn to recover the three types of priors from the current multimodal context. \wbblock{} injects layer-specific combinations of these representations into the action expert through multi-source attention, change- and motion-aware attention biases, and reliability-gated feature modulation. This design grounds action generation in action-relevant future dynamics while adaptively regulating world guidance, enabling robust generalization to visual disturbances and viewpoint shifts, with its generalization performance illustrated in Fig.\ref{fig:banner}(c). 


\vspace{-0.2em}

Compared with direct VLA models~\citep{black2025pi05,zheng2025xvla,ma2026internvla15}, it provides explicit future-aware structure. Compared with WAMs that require large future-model components at training or deployment~\citep{yuan2026fastwam,bi2026motus,ye2026dreamzero,li2026lingbot-va}, it keeps the online path close to a direct action transformer. By emphasizing change and motion rather than raw appearance, the policy is encouraged to ignore visually salient but causally irrelevant scene factors, a property aligned with prior work on robot visual representations and robustness under distribution shift~\citep{nair2022r3m,majumdar2023vip,shah2023voltron,hendrycks2019imagenetc,miller2021manyfaces}. Fig.\ref{fig:banner}(d) presents a comparison of three model architectures. 

Our contributions are summarized as follows:
\vspace{-0.8em}
\begin{itemize}[leftmargin=*, labelsep=0.5em]
    
    \item We propose \method{}, a general world-action framework that learns complementary future-state, spatial-change, and local-motion priors from supervision extracted from observed robot demonstrations. By transferring these priors into compact current-context LWDM, our framework requires neither additional world-model pre-training nor video diffusion transformer execution during policy deployment.

    \item We introduce \wbblock{}, which conditions the action transformer through layer-specific multi-source attention, change- and motion-aware attention biases, and reliability-gated feature modulation. Together with paired clean/augmented training and action-visual-world consistency, this design encourages action-relevant world guidance and stable action generation under observation perturbations.

    \item We demonstrate the generality of \method{} across three VLM backbones, four simulation benchmarks and two real-world robots, achieving consistent out-of-distribution(OOD) gains under appearance and viewpoint shifts and state-of-the-art average success rates on LIBERO-Dynamic, RoboTwin 2.0, and real-world evaluations.

\end{itemize}


\section{Related Work}
\label{sec:relatedwork}

\paragraph{Vision-Language-Action Models.} Vision-language-action (VLA) models connect visual perception and language understanding with robot action generation. Representative models such as RT-2, OpenVLA, $\pi_0$, $\pi_{0.5}$, and InternVLA, among others, advance generalist robot control through pretrained vision-language representations, large-scale robot training, and expressive action architectures~\citep{zitkovich2023rt2,kim2024openvla, black2024pi0,black2025pi05,bjorck2025gr00t,li2024cogact, team2025gemini,cai2026internvla1,ma2026internvla15, cheang2025gr3,bu2025univla,wang2026qwenvla, yuan2026qwenrobotmanip,zheng2025xvla,wu2026linbotvla}. Action chunking and diffusion- or flow-based action modeling further improve the expressiveness and temporal structure of predicted actions~\citep{zhao2023act,chi2023diffusionpolicy, liu2025rdt,black2024pi0,kim2025fine,black2026real}. Despite these advances, many direct VLA models are trained primarily to predict demonstrated actions, with limited explicit supervision for the resulting scene evolution. This can leave models reliant on appearance correlations: changes in background, lighting, or camera viewpoint may disrupt action prediction even when the task and required interaction remain unchanged. Stronger semantic representations and more expressive action decoders do not, by themselves, ensure that models distinguish task-relevant changes from incidental visual variation. \method{} addresses this limitation by guiding action generation with complementary future-state, spatial-change, and local-motion priors, encouraging the model to attend to anticipated task-relevant changes rather than incidental appearance.

\vspace{-0.8em}
\paragraph{World Models for Robotics.}
World models provide predictive representations for planning, policy learning, evaluation, and synthetic data generation in robotics~\citep{hou2026worldmodelsurvey}. Visual foresight, video prediction, and future-generation methods show that anticipated observations can guide manipulation and improve generalization~\citep{finn2017visualforesight,ebert2018visualforesight,babaeizadeh2021fitvid,du2023unipi,zhou2024robodreamer,wu2024gr1,zhou2026gem4d}. Recent world-action models further couple future-state modeling with action generation through latent future reasoning or object-addressable states~\citep{wang2026worldactionmodels,luo2026beingh07,liu2026oawam,yuan2026fastwam}. However, predicting the visual future is not the same as extracting what matters for control: full image/video rollouts can devote substantial computation to reconstructing appearance rather than representing action-relevant changes. Executing a large world model online further burdens every control decision with predictive computation. \method{} instead distills world supervision into compact outcome, change, and motion priors that directly guide action generation, without reconstructing future images or running a large world model at deployment.

\vspace{-0.8em}

\paragraph{Learning Action-Relevant Representations.} A complementary line of work learns explicit spatial and actionable representations for manipulation. Transporter Networks, CLIPort, and Where2Act encode spatial correspondence, separate semantic ``what'' from spatial ``where'', or identify potential interaction regions~\citep{ zeng2021transporter,shridhar2022cliport,mo2021where2act}. PerAct, RVT, and Act3D further exploit 3D structure and multi-view representations for spatially precise action prediction~\citep{shridhar2023peract,goyal2023rvt,gervet2023act3d}. Representation-learning methods such as R3M, VIP, and Voltron investigate visual features that support downstream robot control~\citep{nair2022r3m,majumdar2023vip,shah2023voltron}. Recent world-action models further learn predictive representations through object-addressable dynamics, geometry-aware policy spaces, or compressed world-generator features~\citep{liu2026oawam,zhao2026sgwam,zeng2026enfold}. Unlike these approaches, \method{} explicitly represents complementary future outcomes, spatial changes, and local motion, providing structured world guidance for action generation.


\section{\textsc{Bridge-WA}}
\label{sec:method}
\vspace{-0.4em}
\begin{figure}
    \includegraphics[width=1.02\linewidth]{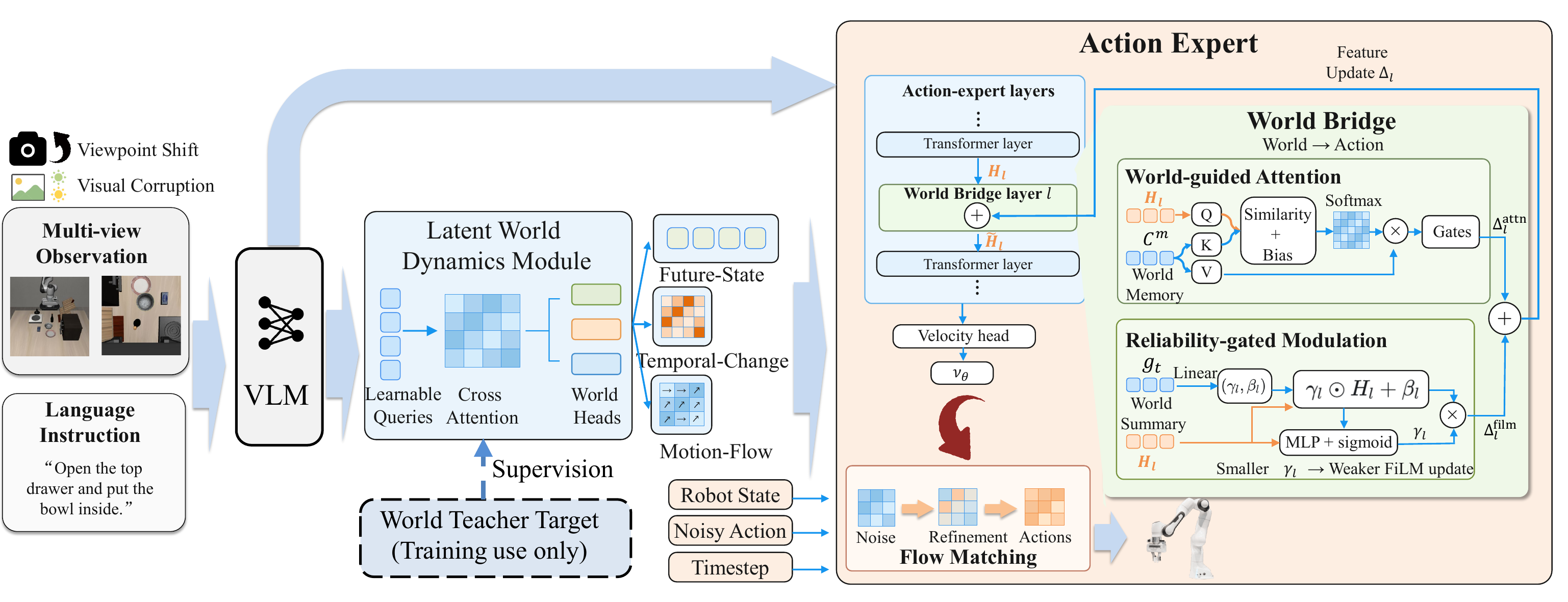}
    \vspace{-18pt}
    \caption{\textbf{The \method{} framework.} A LWDM learns future, change, and flow representations from training-only supervision. \wbblock{}, the world-to-action interface, couples these priors to the action transformer through world-guided attention and reliability-gated FiLM. Flow matching produces the action chunk. Complementary conditioning paths are detailed in Appendix~\ref{app:technical}.}
    \label{fig:bridge_wa_arch}
\end{figure}

\subsection{Overview}
\vspace{-0.4em}
\paragraph{Problem Formulation}
We formulate robotic manipulation as conditional action-sequence generation. Given current multi-view images $I_t=\{I_t^{(v)}\}_{v=1}^{V}$, a language instruction $\ell$, and robot state $s_t$, a policy $\pi_\theta$ generates a chunk of $T_a$ future actions:
\begin{equation}
 A_t=(a_t,\ldots,a_{t+T_a-1})=\pi_\theta(I_t,\ell,s_t),\qquad a_t\in\R^{d_a}.
 \label{eq:policy}
\end{equation}
Here $t$ is the observation timestep, $v$ indexes the $V$ camera views, and $d_a$ is the robot's action dimension; actions control end-effector or joint motion and the gripper. The policy executes a prefix of the chunk before replanning from new observations.
\vspace{-0.8em}
\paragraph{Overview}
\method{} is a general world-action framework that unifies predictive world representation, world-conditioned action generation, and robust policy learning (Fig.~\ref{fig:bridge_wa_arch}). A vision-language backbone first encodes the current context. A latent world predictor then estimates \textbf{what} the scene may become, \textbf{where} it changes, and \textbf{how} local content moves. \wbblock{}, the framework's world-to-action interface, selectively integrates these complementary priors into a flow-matching action transformer. Matched clean/augmented training promotes consistent actions under observation perturbations. In the following sections, we detail the design of: 1) \textbf{Latent World Dynamics Module}, 2) \textbf{\wbblock{}}, and 3) \textbf{Policy Learning}.

\vspace{-0.8em}

\subsection{Latent World Dynamics Module}

\vspace{-0.4em}

\paragraph{Future-World Supervision} We distill complementary future-world information from robot demonstrations into a world predictor. Frozen WAN-VAE and DINOv3 encoders~\citep{wan2025wan,simeoni2025dinov3} extract future tokens $T_f$ from observed future frames. Normalized current-to-future feature differences yield change maps $M_c$, while optical flow provides motion maps $M_o$. Together, these targets describe future outcomes, spatial changes, and local displacements. Targets are cached offline and fused across valid prediction horizons (Appendix~\ref{app:technical}), providing training-only supervision. 

\vspace{-0.8em}

\paragraph{Predictive Representation Distillation} Conditioned on the current multimodal context, five learned-query heads predict WAN/DINO future tokens, WAN/DINO change maps, and motion flow. Regression and cosine-alignment losses form the distillation objective $\mathcal L_W$, transferring future-world information into representations predictable from current inputs. The predictions are organized into three memories: $C^F$ combines predicted WAN/DINO features, $C^C$ fuses the two change predictions, and $C^M$ encodes motion. At inference, the policy uses these current-context representations without future observations or teacher encoders.

\vspace{-0.8em}
\subsection{WorldBridge: Selective World-to-Action Conditioning}
\label{sec:bridge}
\vspace{-0.4em}

\paragraph{Layer-Specific Attention}
Within \method{}, \wbblock{} connects the predicted world representation to action generation. It updates the action-token stream while preserving the multimodal context pathway. At transformer layer $l$, action tokens $H_l$ query each active world memory:
\begin{equation}
 R_l^m=\softmax\!\left(\frac{(H_lW_Q^m)(C^mW_K^m)^\top}{\sqrt{d_h}}
       +\mathbf1(b^m)^\top\right)(C^mW_V^m)W_O^m,
 \quad m\in\{F,C,M\}.
 \label{eq:attention}
\end{equation}
Here $H_l\in\R^{N_a\times d}$ contains the hidden representations of the $N_a$ action tokens entering action-transformer layer $l$, and $d$ is the hidden width. The index $m$ denotes future ($F$), change ($C$), or motion ($M$) memory $C^m$. The matrices $W_Q^m,W_K^m,W_V^m,W_O^m$ are attention projections, $d_h$ is the head width, and $b^m$ is broadcast over action queries. Softmax normalizes over memory tokens, yielding the ungated update $R_l^m$. Future attention uses no extra bias; change scores and motion magnitudes provide key biases for the other two memories. These biases prioritize anticipated changes without imposing explicit temporal positional encodings. Active memories are then gated and applied sequentially:
\begin{equation}
 H_l^{(0)}=H_l,\qquad
 H_l^{(j)}=H_l^{(j-1)}+\Delta_l^{m_j}\!\left(H_l^{(j-1)},C^{m_j}\right),\qquad
 \Delta_l^{\mathrm{attn}}=H_l^{(J)}-H_l.
 \label{eq:attn_residual}
\end{equation}
Here $\mathcal M_l=(m_1,\ldots,m_J)$ is the ordered sequence of the $J$ memories active at layer $l$, and $j$ denotes its position in that sequence. $H_l^{(j)}$ is the action-token stream after the first $j$ memories, initialized by $H_l^{(0)}=H_l$. Each $\Delta_l^{m_j}$ applies modality, layer, channel, and reliability gates to $R_l^{m_j}$; their accumulated change is $\Delta_l^{\mathrm{attn}}$. A layer schedule emphasizes future information earlier and change/motion refinement later. Modalities are applied sequentially, with shared attention weights but distinct layer gates. Each residual is scaled by modality, layer, and channel gates, together with a learned reliability score that compares pooled action and world features. Appendix~\ref{app:technical} specifies the gates and layer schedule.

\vspace{-0.8em}

\paragraph{Reliability-Gated Modulation}
We complement token-level attention with reliability-gated FiLM conditioning~\citep{perez2018film}:
\begin{equation}
 \Delta_l^{\mathrm{film}}=\eta r_l(\gamma_l\odot H_l+\beta_l),\qquad
 \Delta_l=\Delta_l^{\mathrm{attn}}+\Delta_l^{\mathrm{film}},\qquad
 \widetilde H_l=H_l+\Delta_l+\mathcal A(\Delta_l),
 \label{eq:film}
\end{equation}
Here $g_t$ is the fused global world summary(Fig.\ref{fig:bridge_wa_arch}), from which layer-specific channel-wise scale and shift vectors $(\gamma_l,\beta_l)$ are derived. The scalar $\eta$ sets residual strength, and $r_l\in(0,1)$ gates the FiLM residual $\Delta_l^{\mathrm{film}}$ using action--modulation compatibility. The total world update is $\Delta_l=\Delta_l^{\mathrm{attn}}+\Delta_l^{\mathrm{film}}$. The learned adapter $\mathcal A$ further processes this update, producing the conditioned action representation $\widetilde H_l$. Reliability depends on action and modulation features, allowing weaker updates when world guidance is less compatible with the action stream; it is not a calibrated uncertainty estimate. Zero-initialized residual output projections provide a stable initialization for training. Language-conditioned, address-keyed visual slots supply current scene evidence, while bounded state-dependent scalars adjust guidance strength (Appendix~\ref{app:technical}).
\vspace{-0.8em}
\subsection{Robust Policy Learning}
\label{sec:robust}

\vspace{-0.4em}
\paragraph{Matched Clean and Augmented Training}
We train clean and augmented observations of the same demonstration with shared action labels, flow noise $\epsilon$, and interpolation time $\tau$. Mild robot-state, language, and appearance perturbations preserve task supervision. State corruption is aligned between numeric and tokenized channels; appearance perturbations do not relocate objects or modify action labels. A detached, action-gradient-directed visual perturbation additionally regularizes feature sensitivity. Both branches learn the native flow-matching velocity~\citep{lipman2023flow}:
\begin{equation}
 X_\tau=(1-\tau)A_t+\tau\epsilon,\qquad
 \mathcal L_{\mathrm{act}}=\E\|v_\theta(X_\tau,\tau;I_t,\ell,s_t)-(\epsilon-A_t)\|^2.
 \label{eq:fm}
\end{equation}
Here $\epsilon$ is sampled Gaussian noise with the shape of $A_t$, and $\tau\in[0,1]$ is interpolation time, distinct from observation time $t$. $X_\tau$ interpolates from actions to noise; $v_\theta$ predicts its target velocity $\epsilon-A_t$. The expectation averages squared error over demonstrations, noise, and $\tau$. We also align augmented outputs with detached clean targets: SmoothL1 on action velocities and cosine losses on pooled visual features, fused world memories, and visual slots. Denoting their weighted sum by $\mathcal L_{\mathrm{con}}$, the objective is
\begin{equation}
 \mathcal L=\lambda_a\big[(1-\alpha_n)\mathcal L^c_{\mathrm{act}}+
                  \alpha_n\mathcal L^a_{\mathrm{act}}\big]
       +\tfrac12(\mathcal L_W^c+\mathcal L_W^a)+\rho_n\mathcal L_{\mathrm{con}},
 \label{eq:objective}
\end{equation}
Here $c/a$ denote clean/augmented branches, $\lambda_a$ weights action supervision, and $\mathcal L_W$ is world-prior distillation. At optimizer update $n$, $\alpha_n$ balances the two action losses and $\rho_n$ ramps consistency strength; the two world losses are equally weighted. Demonstration reweighting, feature perturbation, and the consistency terms are detailed in Appendix~\ref{app:learning}.

\vspace{-0.8em}

\section{Experiments}

\vspace{-0.8em}
\subsection{Simulation Benchmarks and Real-World Setup}
\vspace{-0.4em}

\begin{figure}
    \centering
    \includegraphics[width=1\linewidth]{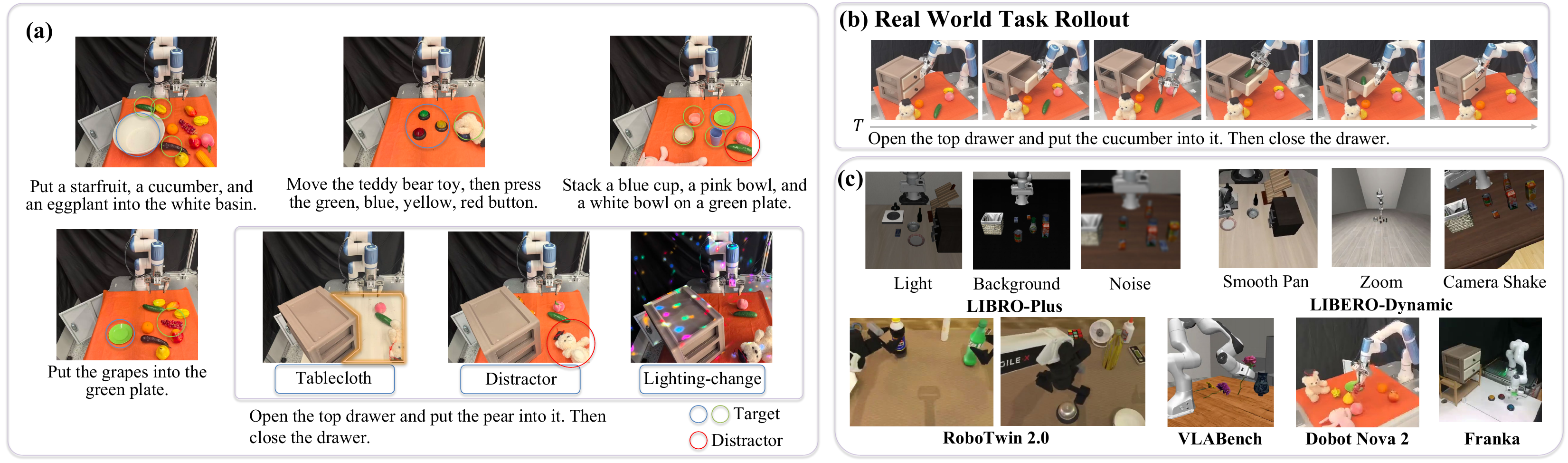}
    \vspace{-18pt}
    \caption{\textbf{Evaluation overview.} (a) Real-world tasks with diverse unseen objects and visual disturbances. (b) A real-world rollout of the ``Put Item in Drawer'' task. (c) Simulation benchmarks and real-world robots used in our evaluation.}
    \label{fig:exp_setup_vis}
\end{figure}

\paragraph{Simulation Benchmarks} We evaluate \method{} on \textbf{LIBERO-Plus}, our \textbf{LIBERO-Dynamic}, \textbf{RoboTwin~2.0}, and \textbf{VLABench} (Fig.~\ref{fig:exp_setup_vis}(c)). Together, these benchmarks test static visual shifts, time-varying viewpoints, bimanual domain randomization, and systematic single-arm generalization. We follow each benchmark's official training and evaluation protocol and report success rate, together with intention and progress scores where defined. Exact datasets, observation and action spaces, action horizons, perturbations, and episode counts are preserved in Appendix~\ref{app:benchmark_protocols}.

\vspace{-1em}

\paragraph{Real-World} We evaluate on the \textbf{Dobot Nova~2} and \textbf{Franka} robots with an Intel RealSense third-person camera and a wrist camera. Demonstrations are stored with RGB images, cartesian robot states, and absolute cartesian actions, with RGB standardized to $640\times480$. The suite contains seven tasks, as shown in the Fig.\ref{fig:exp_setup_vis}(a)(b). Easy evaluation uses 10 in-distribution trials per task; Hard evaluation uses 15 trials per task across tablecloth, distractor-object, and lighting shifts, with five trials per factor. We report per-task and average success rates for each track. More real-world robot experimental setups are provided in the Appendix~\ref{app:real_robot_setup}.

\vspace{-1em}
\paragraph{Experimental Details}
Experimental details, including hardware, VLM backbones, model sizes, and training configurations, are provided in Appendix~\ref{app:experimental_details}.



\vspace{-0.8em}
\subsection{Main Results}
\vspace{-0.4em}
\paragraph{On LIBERO-Plus, LIBERO-Dynamic}

\begin{table*}[t]
    \centering
    \setlength{\tabcolsep}{5.5pt}
    \renewcommand{\arraystretch}{1.15}

    \caption{\textbf{Performance comparison on LIBERO-Plus.} Success rates (\%) across LIBERO-Plus variants and their weighted average. $\Delta$ denotes the gain over the corresponding backbone. $*$ and \textsuperscript{\S} indicate Florence-2 and PaliGemma VLMs, respectively.}
    \vspace{-8pt}
    \definecolor{sectionpink}{RGB}{253,240,240}
    \definecolor{methodblue}{RGB}{238,247,251}
    \definecolor{deltagreen}{RGB}{240,249,238}
    \definecolor{negative}{RGB}{252,235,235}

    \resizebox{\textwidth}{!}{%
    \begin{tabular}{
        >{\centering\arraybackslash}m{5cm}
        *{8}{>{\centering\arraybackslash}m{1.55cm}}
    }
        \toprule
        \textbf{Method}
        & \textbf{Camera}
        & \textbf{Robot}
        & \textbf{Language}
        & \textbf{Light}
        & \textbf{Background}
        & \textbf{Noise}
        & \textbf{Layout}
        & \textbf{Total} \\
        \midrule

        \rowcolor{sectionpink}
        \multicolumn{9}{l}{\textit{World Action Models}} \\

        WorldVLA\citep{cen2025worldvla}
        & 0.1 & 27.9 & 41.6 & 43.7 & 17.1 & 10.9 & 38.0 & 25.0 \\

        Fast-WAM\citep{yuan2026fastwam}
        & 16.4 & 44.5 & 68.9 & 78.2 & 53.7 & 37.7 & 60.7 & 51.5 \\

        SG-WAM\citep{zhao2026sgwam}
        & 58.6 & 48.9 & 81.4 & 89.8 & 86.1 & 80.7 & 74.2 & 73.0 \\

        Faster-WAM\citep{zhao2026faster1}
        & 53.8 & 71.6 & \textbf{94.7} & 96.3 & 61.3 & 63.6 & 79.1 & 73.6 \\

        Faster-WAM\citep{ma2026faster2}
        & 67.9 & 49.0 & 92.1 & 94.3 & 57.0 & 82.3 & 82.7 & 75.0 \\

        VLA-JEPA\citep{sun2026vlajepa}
        & 64.2 & 67.7 & 88.1 & 91.8 & 93.4 & 65.8 & 83.9 & 79.5 \\

        GE-Act\citep{liao2026geact}
        & 60.7 & 77.0 & 77.4 & 95.8 & 86.0 & 90.9 & 80.2 & 80.3 \\

        Cosmos-Policy\citep{kim2026cosmos}
        & 75.8 & 63.3 & 81.7 & 96.5 & 88.9 & 92.7 & 82.2 & 82.2 \\

        ImageWAM\citep{zhang2026imagewam}
        & 80.8 & 50.3 & 91.4 & 98.1 & 85.5 & 93.8 & 80.5 & 83.1 \\

        OA-WAM\citep{liu2026oawam}
        & 80.5 & 89.6 & 85.3 & 96.5 & 95.9 & 75.6 & 82.8 & 83.9 \\

        \midrule
        \rowcolor{sectionpink}
        \multicolumn{9}{l}{\textit{Vision-Language-Action Models}} \\

        OpenVLA\citep{kim2024openvla}
        & 0.8 & 3.5 & 23.0 & 8.1 & 34.8 & 15.2 & 28.5 & 15.6 \\

        NORA\citep{hung2025nora}
        & 2.2 & 37.0 & 65.1 & 45.7 & 58.6 & 12.8 & 62.1 & 39.0 \\

        UniVLA\citep{bu2025univla}
        & 1.8 & 46.2 & 69.6 & 69.0 & 81.0 & 21.2 & 31.9 & 42.9 \\

        $\pi_{0}$\cite{black2024pi0}
        & 13.8 & 6.0 & 58.8 & 85.0 & 81.4 & 79.0 & 68.9 & 53.6 \\


        $\pi_{0}$-Fast\citep{pertsch2025pi0fast}
        & 65.1 & 21.6 & 61.0 & 73.2 & 73.2 & 74.4 & 68.8 & 61.6 \\


        RIPT-VLA\citep{tan2025RIPT-VLA}
        & 55.2 & 31.2 & 77.6 & 88.4 & 91.6 & 73.5 & 74.2 & 68.4 \\

        OpenVLA-OFT\citep{kim2025fine}
        & 56.4 & 31.9 & 79.5 & 88.7 & 93.3 & 75.8 & 74.2 & 69.6 \\

        AVA-VLA\citep{xiao2026ava}
        & 55.5 & 25.9 & 85.6 & 95.5 & 88.9 & 78.0 & 74.1 & 70.1 \\

        HoloBrain-0\citep{lin2026holobrain}
        & 65.5 & 58.2 & 78.7 & 88.1 & 90.3 & 66.9 & 79.5 & 74.0 \\

        StarVLA\citep{community2026starvla}
        & 52.5 & 49.8 & 88.5 & 95.7 & 95.7 & 73.0 & 76.9 & 74.1 \\

        ABot-M0\citep{yang2026abot}
        & 60.4 & 67.9 & 86.4 & 96.2 & 91.6 & 86.4 & 82.6 & 80.5 \\

        \bottomrule
        \addlinespace[2pt]
        \rowcolor{methodblue}
        X-VLA\citep{zheng2025xvla}
        & 23.4 & 89.7 & 75.7 & 88.2 & 96.0 & 62.7 & 71.8 & 71.4 \\

        \rowcolor{methodblue}
        \method{}\textsuperscript{$*$} (Ours)
        & 60.7 & \textbf{90.6} & 81.0 & 98.6 & 95.5 & 86.9 & 72.5 & 82.5 \\

        \rowcolor{deltagreen}
        $\Delta$
        & $+37.3$
        & $+0.9$
        & $+5.3$
        & $+10.4$
        & \cellcolor{negative}$-0.5$
        & $+24.2$
        & $+0.7$
        & $+11.1$ \\

        \addlinespace[2pt]
        \rowcolor{methodblue}
        $\pi_{0.5}$\citep{black2025pi05}
        & 75.4 & 77.5 & 85.6 & 96.9 & 94.6 & 89.7 & 85.7 & 85.7 \\

        \rowcolor{methodblue}
        \method{}\textsuperscript{\S} (Ours)
        & 77.2 & 85.6 & 86.7 & 98.3 & 97.9 & 92.5 & \textbf{88.5} & \textbf{88.7} \\

        \rowcolor{deltagreen}
        $\Delta$
        & $+1.8$
        & $+8.1$
        & $+1.1$
        & $+1.3$
        & $+3.3$
        & $+2.8$
        & $+2.8$
        & $+3.0$ \\

        \addlinespace[2pt]
        \rowcolor{methodblue}
        InternVLA-A1.5\citep{ma2026internvla15}
        & 83.1 & 55.1 & 86.9 & 96.4 & 98.2 & 95.6 & 85.2 & 84.8 \\

        \rowcolor{methodblue}
        \method{} (Ours)
        & \textbf{89.8} & 67.0 & 85.7 & \textbf{98.7} & \textbf{98.5} & \textbf{98.5} & 83.6 & 88.0 \\

        \rowcolor{deltagreen}
        $\Delta$
        & $+6.7$
        & $+11.9$
        & \cellcolor{negative}$-1.2$
        & $+2.3$
        & $+0.3$
        & $+2.9$
        & \cellcolor{negative}$-1.6$
        & $+3.2$ \\

        \bottomrule
    \end{tabular}%
    }
    
    \label{tab:libero_plus}
\end{table*}

\begin{table*}[t]
\centering
\footnotesize
\setlength{\tabcolsep}{6pt}
\renewcommand{\arraystretch}{1.12}

\definecolor{methodblue}{RGB}{238,247,251}
\definecolor{deltagreen}{RGB}{240,249,238}

\caption{\textbf{Performance comparison on LIBERO-Dynamic.} Success rates (\%) across LIBERO-Dynamic camera-motion profiles and their average.}
\vspace{-8pt}
\resizebox{0.8\textwidth}{!}{%
\begin{tabular}{
    >{\centering\arraybackslash}m{5cm}
    *{6}{>{\centering\arraybackslash}m{1.65cm}}
}
    \toprule

    \textbf{Method}
    & \textbf{Smooth Pan}
    & \textbf{Orbit}
    & \textbf{Zoom}
    & \textbf{View Jump}
    & \textbf{Camera Shake}
    & \textbf{Total} \\
    \midrule

    \rowcolor{methodblue}
    X-VLA\citep{zheng2025xvla}
    & 37.65
    & 38.40
    & 27.45
    & 21.30
    & 46.95
    & 34.35 \\

    \rowcolor{methodblue}
    \method{}$^*$ (Ours)
    & 81.65
    & 77.80
    & 70.25
    & 72.90
    & 79.20
    & 76.36 \\

    \rowcolor{deltagreen}
    $\Delta$
    & $+44.00$
    & $+39.40$
    & $+42.80$
    & $+51.60$
    & $+32.25$
    & $+42.01$ \\


    \rowcolor{methodblue}
    InternVLA-A1.5\citep{ma2026internvla15}
    & 93.10
    & 93.30
    & 92.35
    & 93.35
    & 95.15
    & 93.45 \\

    \rowcolor{methodblue}
    \method{} (Ours)
    & \textbf{93.20}
    & \textbf{93.85}
    & \textbf{93.90}
    & \textbf{93.50}
    & \textbf{96.05}
    & \textbf{94.10} \\

    \rowcolor{deltagreen}
    $\Delta$
    & $+0.10$
    & $+0.55$
    & $+1.55$
    & $+0.15$
    & $+0.90$
    & $+0.65$ \\

    \bottomrule
\end{tabular}%
}

\label{tab:libero_dynamic}
\end{table*}

As shown in Tables~\ref{tab:libero_plus} and~\ref{tab:libero_dynamic}, \method{} consistently improves three heterogeneous VLA backbones—X-VLA, $\pi_{0.5}$, and InternVLA-A1.5—built on Florence-2, PaliGemma, and Qwen3.5, respectively. 

\begin{wraptable}{r}{0.50\textwidth}
    \centering
    \vspace{-4pt}

    \setlength{\tabcolsep}{7pt}
    \renewcommand{\arraystretch}{1.15}

    \caption{
        \textbf{Performance on RoboTwin 2.0.}
        Success rates (\%) on the Easy and Hard settings of
        RoboTwin~2.0 and their average.
    }
    \vspace{-8pt}
    \resizebox{0.8\linewidth}{!}{%
        \begin{tabular}{
            >{\centering\arraybackslash}m{5.5cm}
            *{3}{>{\centering\arraybackslash}m{1.2cm}}
        }
            \toprule
            \textbf{Method}
            & \textbf{Easy}
            & \textbf{Hard}
            & \textbf{Average} \\
            \midrule

            \rowcolor{sectionpink}
            \multicolumn{4}{l}{\textit{World Action Models}} \\

            Motus\citep{bi2026motus}
            & 88.7 & 87.0 & 87.8 \\

            Faster-WAM\citep{ma2026faster2}
            & 89.7 & 88.6 & 89.2 \\

            Being-H0.7\citep{luo2026beingh07}
            & 90.2 & 89.6 & 89.9 \\

            Fast-WAM\citep{yuan2026fastwam}
            & 91.9 & 91.8 & 91.8 \\

            LingBot-VA\citep{li2026lingbot-va}
            & 92.9 & 91.5 & 92.2 \\

            Enfold\citep{zeng2026enfold}
            & 91.6 & 91.9 & 91.8 \\

            \midrule
            \rowcolor{sectionpink}
            \multicolumn{4}{l}{%
                \textit{Vision-Language-Action Models}
            } \\

            $\pi_{0}$\citep{black2024pi0}
            & 65.9 & 58.4 & 62.2 \\

            StarVLA\citep{community2026starvla}
            & 88.7 & 87.8 & 88.3 \\

            $\pi_{0.5}$\citep{black2025pi05}
            & 82.7 & 76.8 & 79.8 \\

            LingBot-VLA\citep{wu2026linbotvla}
            & 86.5 & 85.3 & 85.9 \\

            InternVLA-A1\citep{cai2026internvla1}
            & 89.4 & 89.6 & 89.5 \\

            \midrule

            \rowcolor{methodblue}
            X-VLA\citep{zheng2025xvla}
            & 72.9 & 72.8 & 72.9 \\

            \rowcolor{methodblue}
            \method{}\textsuperscript{*} (Ours)
            & 75.2 & 75.7 & 75.4 \\

            \rowcolor{deltagreen}
            $\Delta$
            & $+2.3$ & $+2.9$ & $+2.6$ \\

            \addlinespace[2pt]

            \rowcolor{methodblue}
            InternVLA-A1.5\citep{ma2026internvla15}
            & 93.3 & 93.0 & 93.2 \\

            \rowcolor{methodblue}
            \method{} (Ours)
            & \textbf{97.1}
            & \textbf{96.6}
            & \textbf{96.9} \\

            \rowcolor{deltagreen}
            $\Delta$
            & $+3.8$ & $+3.6$ & $+3.7$ \\

            \bottomrule
        \end{tabular}%
    }

    \label{tab:robotwin2}
    \vspace{-8pt}
\end{wraptable}

On LIBERO-Plus, it raises their weighted success rates by $11.1$, $3.0$, and $3.2$ points, achieving a best overall score of $88.7$ without using any LIBERO-Plus data. The largest gains occur under camera, robot-state, and visual-noise shifts, where policies must distinguish task-relevant changes from nuisance observations. On LIBERO-Dynamic, \method{} improves X-VLA by $42.01$ points and further raises the strong InternVLA-A1.5 baseline across every camera-motion profile. These consistent gains across different VLM scales and policy architectures demonstrate that \method{} provides a general world-to-action interface rather than a backbone-specific modification.

\vspace{-1.2em}

\paragraph{On RoboTwin~2.0}

Table~\ref{tab:robotwin2} shows that \method{} consistently improves language-conditioned bimanual manipulation. It improves X-VLA by $2.6$ points and InternVLA-A1.5 by $3.7$ points, reaching $97.1/96.6$ success rates on the Easy/Hard settings and a state-of-the-art average of $96.9$. Importantly, the gain is preserved under the Hard setting, which jointly varies distractors, textures, lighting, tabletop height, and instructions. The improvement on a bimanual platform with 14-dimensional actions and a 50-step action horizon indicates that structured world priors extend beyond single-arm, short-horizon control.

\vspace{-1.2em}
\paragraph{On VLABench}

\begin{table*}[t]
    \centering
    
    \setlength{\tabcolsep}{3.5pt}
    \renewcommand{\arraystretch}{1.15}

    \caption{\textbf{Performance comparison on VLABench.}}
    \vspace{-8pt}
    \definecolor{methodblue}{RGB}{238,247,251}
    \definecolor{deltagreen}{RGB}{240,249,238}
    \definecolor{negative}{RGB}{252,235,235}

    \resizebox{\textwidth}{!}{%
    \begin{tabular}{
        >{\centering\arraybackslash}m{4cm}
        *{18}{>{\centering\arraybackslash}m{0.78cm}}
    }
        \toprule

        \multirow{2}{*}{\textbf{Method}}
        & \multicolumn{3}{c}{\textbf{In-Dist.}}
        & \multicolumn{3}{c}{\textbf{Cross-Cat.}}
        & \multicolumn{3}{c}{\textbf{Common Sense}}
        & \multicolumn{3}{c}{\textbf{Semantic Instr.}}
        & \multicolumn{3}{c}{\textbf{Unseen Texture}}
        & \multicolumn{3}{c}{\textbf{Avg.}} \\

        \cmidrule(lr){2-4}
        \cmidrule(lr){5-7}
        \cmidrule(lr){8-10}
        \cmidrule(lr){11-13}
        \cmidrule(lr){14-16}
        \cmidrule(lr){17-19}

        & \textbf{SR} & \textbf{IS} & \textbf{PS}
        & \textbf{SR} & \textbf{IS} & \textbf{PS}
        & \textbf{SR} & \textbf{IS} & \textbf{PS}
        & \textbf{SR} & \textbf{IS} & \textbf{PS}
        & \textbf{SR} & \textbf{IS} & \textbf{PS}
        & \textbf{SR} & \textbf{IS} & \textbf{PS} \\
        \midrule

        $\pi_{0.5}$ \citep{black2025pi05}
        & 40.6 & 75.0 & 60.8
        & 22.6 & 49.6 & 35.3
        & 18.0 & 57.5 & 41.6
        & 16.1 & 57.1 & 30.3
        & 25.6 & 62.0 & 47.4
        & 24.6 & 60.2 & 43.1 \\

        $\pi_{0}$\citep{black2024pi0}
        & 47.0 & 67.8 & 62.7
        & 21.2 & 44.0 & 33.6
        & 29.1 & 54.9 & 43.0
        & 17.3 & 58.0 & 38.7
        & 32.2 & 50.6 & 42.5
        & 29.4 & 55.0 & 44.1 \\

        \addlinespace[2pt]
        \rowcolor{methodblue}
        X-VLA\citep{zheng2025xvla}
        & 40.0 & 88.0 & 66.5
        & 11.0 & 42.0 & 24.0
        & 33.0 & 66.0 & 51.7
        & 39.0 & 82.0 & 62.5
        & 24.0 & 73.0 & 51.2
        & 29.4 & 70.2 & 51.2 \\

        \rowcolor{methodblue}
        \method{}\textsuperscript{*} (Ours)
        & 78.0 & 85.0 & 85.8
        & 23.0 & 39.0 & 28.8
        & 51.1 & 74.2 & 64.4
        & 67.0 & 82.0 & 80.3
        & 45.0 & 76.0 & 60.3
        & 52.8 & 71.2 & 64.0 \\

        \rowcolor{deltagreen}
        $\Delta$
        & $+38.0$
        & \cellcolor{negative}$-3.0$
        & $+19.3$
        & $+12.0$
        & \cellcolor{negative}$-3.0$
        & $+4.8$
        & $+18.1$
        & $+8.2$
        & $+12.7$
        & $+28.0$
        & $+0.0$
        & $+17.8$
        & $+21.0$
        & $+3.0$
        & $+9.1$
        & $+23.4$
        & $+1.0$
        & $+12.8$ \\




        \bottomrule
    \end{tabular}%
    }
    
    \label{tab:vlabench_results}
\end{table*}

\begin{table*}[t]
    \centering
    \small
    \setlength{\tabcolsep}{3pt}
    \renewcommand{\arraystretch}{1.15}

     \caption{
        \textbf{Component masking ablation on VLABench.} F, C, and O denote future, change, and flow priors, respectively.
    }
    \vspace{-8pt}
    \definecolor{methodblue}{RGB}{238,247,251}

    \resizebox{0.7\textwidth}{!}{%
    \begin{tabular}{
        *{3}{>{\centering\arraybackslash}m{0.55cm}}
        *{12}{>{\centering\arraybackslash}m{0.72cm}}
    }
        \toprule

        \multicolumn{3}{c}{\textbf{World Prior}}
        & \multicolumn{2}{c}{\textbf{In-Dist.}}
        & \multicolumn{2}{c}{\textbf{Cross-Cat.}}
        & \multicolumn{2}{c}{\shortstack{\textbf{Common}\\\textbf{Sense}}}
        & \multicolumn{2}{c}{\shortstack{\textbf{Semantic}\\\textbf{Instr.}}}
        & \multicolumn{2}{c}{\shortstack{\textbf{Unseen}\\\textbf{Texture}}}
        & \multicolumn{2}{c}{\textbf{Avg.}} \\

        \cmidrule(lr){1-3}
        \cmidrule(lr){4-5}
        \cmidrule(lr){6-7}
        \cmidrule(lr){8-9}
        \cmidrule(lr){10-11}
        \cmidrule(lr){12-13}
        \cmidrule(lr){14-15}

        \textbf{F} & \textbf{C} & \textbf{O}
        & \textbf{SR} & \textbf{PS}
        & \textbf{SR} & \textbf{PS}
        & \textbf{SR} & \textbf{PS}
        & \textbf{SR} & \textbf{PS}
        & \textbf{SR} & \textbf{PS}
        & \textbf{SR} & \textbf{PS} \\
        \midrule

        & &
        & 17.0 & 51.0
        & 12.0 & 28.2
        & 14.0 & 47.3
        & 15.0 & 48.7
        & 15.0 & 45.2
        & 14.6 & 44.1 \\

        $\checkmark$ & &
        & 73.0 & 82.5
        & 18.0 & 24.3
        & 43.0 & 57.0
        & 60.0 & 74.3
        & 42.0 & 57.7
        & 47.2 & 59.2 \\

        & $\checkmark$ &
        & 34.0 & 63.3
        & 19.0 & 31.2
        & 27.0 & 53.3
        & 34.0 & 61.8
        & 32.0 & 53.7
        & 29.2 & 52.7 \\

        & & $\checkmark$
        & 20.0 & 31.7
        & 4.0 & 10.8
        & 8.0 & 18.7
        & 22.0 & 30.0
        & 7.0 & 18.5
        & 12.2 & 21.9 \\

        $\checkmark$ & $\checkmark$ &
        & 76.0 & 85.0
        & \textbf{25.0} & \textbf{31.5}
        & 46.0 & 61.7
        & 65.0 & 78.7
        & 43.0 & 61.2
        & 51.0 & 63.6 \\

        & $\checkmark$ & $\checkmark$
        & 45.0 & 66.3
        & 6.0 & 16.8
        & 24.0 & 44.5
        & 38.0 & 58.8
        & 26.0 & 44.3
        & 27.8 & 46.2 \\

        $\checkmark$ & & $\checkmark$
        & 77.0 & \textbf{86.3}
        & 21.0 & 28.0
        & 47.0 & 60.8
        & 66.0 & 79.8
        & 42.0 & 59.3
        & 50.6 & 62.9 \\

        \rowcolor{methodblue}
        $\checkmark$ & $\checkmark$ & $\checkmark$
        & \textbf{78.0} & 85.8
        & 23.0 & 28.8
        & \textbf{51.1} & \textbf{64.4}
        & \textbf{67.0} & \textbf{80.3}
        & \textbf{45.0} & \textbf{60.3}
        & \textbf{52.8} & \textbf{64.0} \\

        \bottomrule
    \end{tabular}%
    }
    
    \label{tab:vlabench_component_ablation}
\end{table*}

As reported in Table~\ref{tab:vlabench_results}, \method{} substantially improves both task completion and execution progress across all five VLABench tracks. Average SR increases from $29.4$ to $52.8$ and PS from $51.2$ to $64.0$, with particularly large SR gains on In-Distribution ($+38.0$), Semantic Instruction ($+28.0$), and Unseen Texture ($+21.0$). In contrast, the average intention score changes only marginally. This suggests that the improvement does not primarily arise from better instruction recognition, but from translating semantic intent into more effective actions. By jointly representing the intended future, spatial support of change, and local motion, \method{} improves execution under semantic, categorical, and appearance shifts.

\vspace{-1.2em}

\begin{figure}[!t]
    \centering
    \includegraphics[width=1\linewidth]{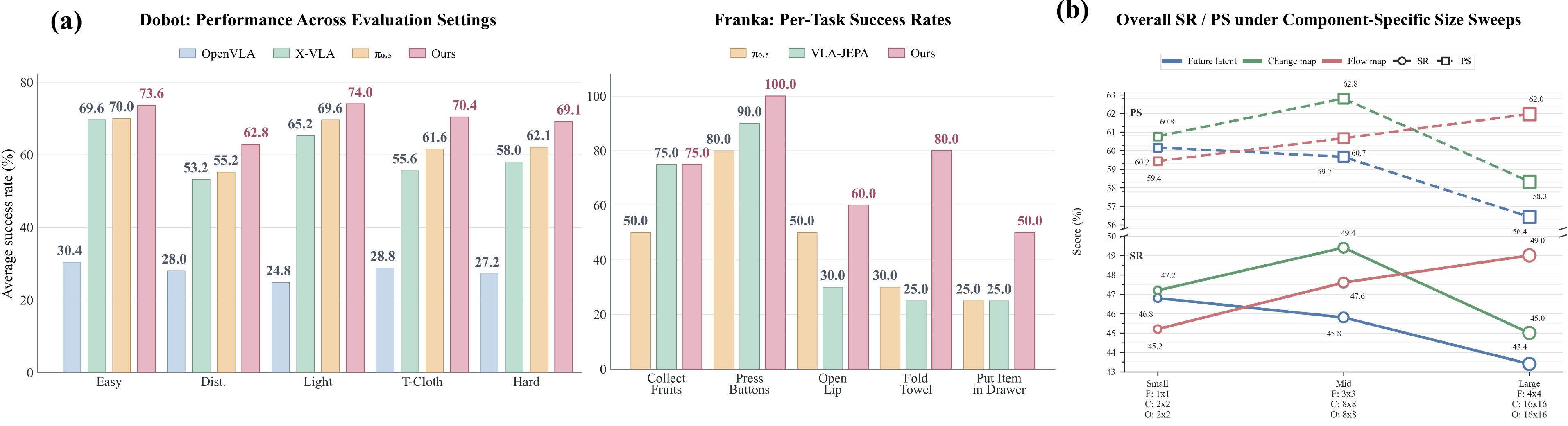}
    \vspace{-20pt}
    \caption{\textbf{Real-world evaluation and component analysis.} (a) Performance across visual disturbances on Dobot and per-task success rates on Franka. The detailed success rates on Dobot robot are reported in the Appendix~\ref{app:additional_experimental_results}. (b) Overall SR and PS under different resolutions of the future, change, and flow representations on VLABench.}
    \label{fig:dobot-franka-size}
\end{figure}

\paragraph{On Real-World Robots}
Fig.\ref{fig:dobot-franka-size}(a) show that the simulation gains transfer to both Dobot Nova~2 and Franka robots. On Dobot, \method{} improves the average Easy success rate from $69.6$ to $73.6$ and the aggregated Hard result from $58.0$ to $69.1$, with the larger OOD gain indicating improved robustness to distractors, lighting, and tablecloth changes. On Franka, it achieves a $73.0$ average success rate, compared with $49.0$ for the strongest baseline. These gains are obtained with only 50 demonstrations per task, supporting the claim that action-relevant world priors suppress nuisance appearance while preserving the scene dynamics required for robust execution.

\vspace{-0.8em}
\subsection{Ablation Studies}
\vspace{-0.4em}

\paragraph{World Component Analysis}
Table~\ref{tab:vlabench_component_ablation} reveals complementary but asymmetric roles of the three world priors. Future representations provide the strongest individual contribution, improving average SR/PS from $14.6/44.1$ to $47.2/59.2$, confirming that anticipated outcomes are the primary guidance for action generation. Change maps also yield clear gains by localizing where interactions should occur, whereas flow alone is insufficient because local motion is ambiguous without task outcome and spatial context. When combined with future and change priors, however, flow provides additional motion guidance. The full model achieves the best overall SR/PS of $52.8/64.0$ and the strongest results on Common Sense, Semantic Instruction, and Unseen Texture. These results support our ``what--where--how'' decomposition: future priors define the goal, change priors localize its spatial support, and flow priors refine how the change should unfold.


\vspace{-1em}

\paragraph{WorldBridge Architecture Analysis}

\begin{wraptable}{r}{0.32\textwidth}
    \centering
    
    \caption{\textbf{\wbblock{} Ablation results on LIBERO-Plus.}}
    \vspace{-8pt}
    \setlength{\tabcolsep}{10pt}
    \renewcommand{\arraystretch}{1.10}

    \resizebox{\linewidth}{!}{%
        \begin{tabular}{lc}
            \toprule
            \textbf{Config} & \textbf{Total} \\
            \midrule

            \rowcolor{cyan!12}
            \textbf{Full Model} & \textbf{82.45} \\

            \midrule
            w/o Multi-Source Attention
                & 81.28 \\
            w/o Spatiotemporal Attention Bias
                & 82.13 \\
            w/o Modulation Reliability Gate
                & 75.87 \\
            w/o Visual Invariance Adapter
                & 82.24 \\
            w/o All Reliability Gates
                & 71.34 \\
            \bottomrule
        \end{tabular}%
    }

    \label{tab:libero_plus_ablation}
    \vspace{-4pt}
\end{wraptable}

Table~\ref{tab:libero_plus_ablation} shows that WorldBridge benefits primarily from reliability-aware integration rather than unregulated feature injection. Removing the modulation reliability gate reduces performance by $6.58$ points, while removing reliability gating from both guided attention and feature modulation causes the largest drop of $11.11$ points. This confirms that predicted world priors have input-dependent reliability and must be adaptively regulated before influencing action generation. Multi-source attention contributes $1.17$ points by allowing the action transformer to access complementary world memories separately. The smaller drops from removing the spatiotemporal attention bias or visual invariance adapter suggest that their functions are partially supported by content-based attention and robust training, while their inclusion still improves the complete model. Overall, reliability gating is the central mechanism that converts predictive world features into stable action guidance.

\vspace{-1em}

\paragraph{Future-Prediction Transfer Analysis}
WorldBridge captures predictive information beyond current VLM features. Diagnostic heads trained on LIBERO and evaluated zero-shot on LIBERO-Plus show consistently lower changed-region prediction error across future horizons, including a $2.21\%$ relative reduction at horizon $+30$. Positive improvements across all perturbation families demonstrate transferable future-state modeling under distribution shifts; full results are provided in Appendix~\ref{app:Future-Prediction-Transfer-Analysis}.

\vspace{-1em}

\paragraph{World-prior Size Sweep Analysis} The size sweep in Fig.~\ref{fig:dobot-franka-size}(b) shows that each prior needs a different spatial resolution budget. Future latents work best as compact global summaries: 1$\times$1 reaches 46.8\% SR and 60.2\% PS, while 4$\times$4 drops to 43.4\% SR and 56.4\% PS. Change maps peak at the intermediate 8$\times$8 scale (49.4\% SR / 62.8\% PS), balancing localization and stability. Flow maps benefit from finer resolution, improving from 45.2\% SR at 2$\times$2 to 49.0\% SR at 16$\times$16. Overall, \method{} benefits from asymmetric capacity allocation: compact future tokens, medium-resolution change maps, and higher-resolution flow maps.

\begin{figure}[!t]
    \centering
    \includegraphics[width=1\linewidth]{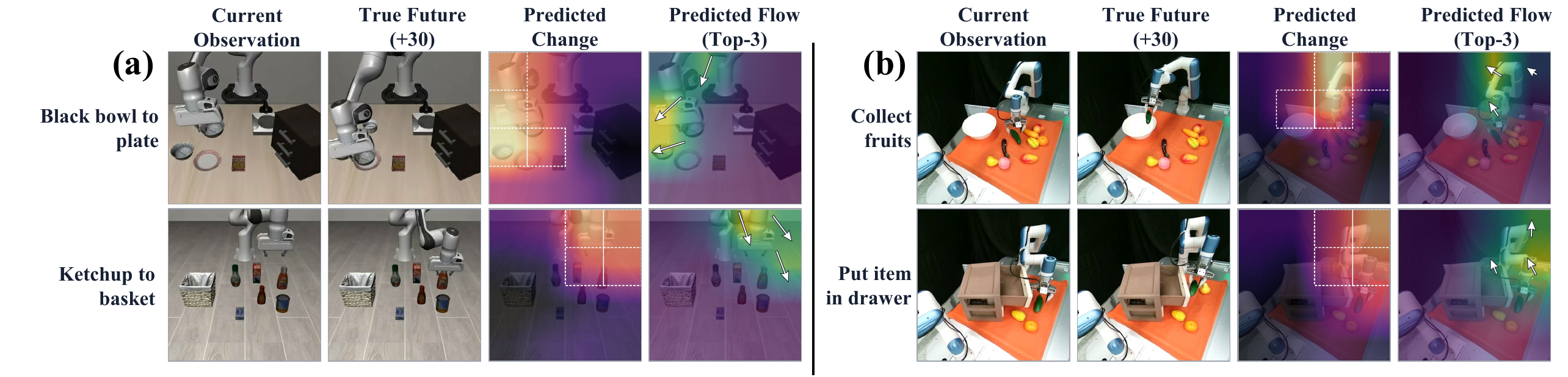}
    \vspace{-18pt}
    \caption{\textbf{Spatial interpretation.} Given the current observation, \method{} predicts action-relevant change regions and the dominant 2-D motion vectors. The predicted changes concentrate on the robot and manipulated objects, while the flow directions are consistent with the observed future motion.}
    \label{fig:spatial_interpretation}
\end{figure}

\vspace{-0.8em}
\subsection{Visualization Analysis}
\vspace{-0.8em}
We visualize the world priors learned by \method{} in both simulation and real-world environments. As shown in Fig.~\ref{fig:spatial_interpretation}(a)(b), the change maps identify \textbf{where} action-relevant changes occur, while suppressing backgrounds and distractors; the flow vectors capture \textbf{how} manipulated regions move and align with the observed future motion. Together with future-state representations of \textbf{what} the scene may become, these priors focus action generation on relevant dynamics, improving robustness to visual and viewpoint shifts.

\vspace{-0.8em}
\section{Conclusion and Limitations}
\label{sec:conclusion}
\vspace{-0.8em}

We presented \method{}, a general world-action framework that replaces dense future-scene generation with action-relevant representations of \textbf{what} may happen, \textbf{where} changes occur, and \textbf{how} they unfold. LWDM predicts these priors from the current context, while \wbblock{} selectively injects them into the action transformer through reliability-aware conditioning. Across three VLA backbones, four simulation benchmarks, and two real-robot platforms, \method{} consistently improves manipulation robustness, particularly under visual and viewpoint shifts. However, the ablations further confirm the complementarity of the world priors and the importance of reliability gating. The learned priors depend on demonstration coverage and may not fully capture occluded 3D dynamics. Future work will explore geometry-aware priors and broader cross-embodiment training.

\clearpage

\bibliography{iclr2027_conference}

@inproceedings{zitkovich2023rt2,
  title={Rt-2: Vision-language-action models transfer web knowledge to robotic control},
  author={Zitkovich, Brianna and Yu, Tianhe and Xu, Sichun and Xu, Peng and Xiao, Ted and Xia, Fei and Wu, Jialin and Wohlhart, Paul and Welker, Stefan and Wahid, Ayzaan and others},
  booktitle={Conference on Robot Learning},
  pages={2165--2183},
  year={2023},
  organization={PMLR}
}

@inproceedings{oneill2024openx,
  title={Open x-embodiment: Robotic learning datasets and rt-x models: Open x-embodiment collaboration 0},
  author={O’Neill, Abby and Rehman, Abdul and Maddukuri, Abhiram and Gupta, Abhishek and Padalkar, Abhishek and Lee, Abraham and Pooley, Acorn and Gupta, Agrim and Mandlekar, Ajay and Jain, Ajinkya and others},
  booktitle={2024 IEEE International Conference on Robotics and Automation (ICRA)},
  pages={6892--6903},
  year={2024},
  organization={IEEE}
}

@inproceedings{kim2024openvla,
  title={{OpenVLA}: An Open-Source Vision-Language-Action Model},
  author={Kim, Moo Jin and Pertsch, Karl and Karamcheti, Siddharth and Xiao, Ted and Balakrishna, Ashwin and Nair, Suraj and Rafailov, Rafael and Foster, Ethan P and Sanketi, Pannag R and Vuong, Quan and others},
  booktitle={8th Annual Conference on Robot Learning},
  year={2024}
}

@article{octo2024,
  title={Octo: An open-source generalist robot policy},
  author={Team, Octo Model and Ghosh, Dibya and Walke, Homer and Pertsch, Karl and Black, Kevin and Mees, Oier and Dasari, Sudeep and Hejna, Joey and Kreiman, Tobias and Xu, Charles and others},
  journal={arXiv preprint arXiv:2405.12213},
  year={2024}
}

@article{black2024pi0,
  title={Pi0: A Vision-Language-Action Flow Model for General Robot Control},
  author={Black, Kevin and Brown, Noah and Driess, Danny and Esmail, Adnan and Equi, Michael and Finn, Chelsea and Fusai, Niccolo and Groom, Lachy and Hausman, Karol and Ichter, Brian and others},
  journal={arXiv preprint arXiv:2410.24164},
  year={2024}
}

@article{chi2023diffusionpolicy,
  title={Diffusion policy: Visuomotor policy learning via action diffusion},
  author={Chi, Cheng and Xu, Zhenjia and Feng, Siyuan and Cousineau, Eric and Du, Yilun and Burchfiel, Benjamin and Tedrake, Russ and Song, Shuran},
  journal={The International Journal of Robotics Research},
  volume={44},
  number={10-11},
  pages={1684--1704},
  year={2025},
  publisher={Sage Publications Sage UK: London, England}
}

@article{du2023unipi,
  title={Learning universal policies via text-guided video generation},
  author={Du, Yilun and Yang, Sherry and Dai, Bo and Dai, Hanjun and Nachum, Ofir and Tenenbaum, Josh and Schuurmans, Dale and Abbeel, Pieter},
  journal={Advances in neural information processing systems},
  volume={36},
  pages={9156--9172},
  year={2023}
}

@inproceedings{zhou2024robodreamer,
  title={RoboDreamer: Learning Compositional World Models for Robot Imagination},
  author={Zhou, Siyuan and Du, Yilun and Chen, Jiaben and Li, Yandong and Yeung, Dit-Yan and Gan, Chuang},
  booktitle={International Conference on Machine Learning},
  pages={61885--61896},
  year={2024},
  organization={PMLR}
}

@article{liu2023libero,
  title={Libero: Benchmarking knowledge transfer for lifelong robot learning},
  author={Liu, Bo and Zhu, Yifeng and Gao, Chongkai and Feng, Yihao and Liu, Qiang and Zhu, Yuke and Stone, Peter},
  journal={Advances in Neural Information Processing Systems},
  volume={36},
  pages={44776--44791},
  year={2023}
}

@article{zhang2024vlabench,
  title={Vlabench: A large-scale benchmark for language-conditioned robotics manipulation with long-horizon reasoning tasks, 2024a},
  author={Zhang, Shiduo and Xu, Zhe and Liu, Peiju and Yu, Xiaopeng and Li, Yuan and Gao, Qinghui and Fei, Zhaoye and Yin, Zhangyue and Wu, Zuxuan and Jiang, Yu-Gang and others},
  journal={URL https://arxiv. org/abs/2412.18194}
}

@article{chen2025robotwin2,
  title={Robotwin 2.0: A scalable data generator and benchmark with strong domain randomization for robust bimanual robotic manipulation},
  author={Chen, Tianxing and Chen, Zanxin and Chen, Baijun and Cai, Zijian and Liu, Yibin and Li, Zixuan and Liang, Qiwei and Lin, Xianliang and Ge, Yiheng and Gu, Zhenyu and others},
  journal={arXiv preprint arXiv:2506.18088},
  year={2025}
}

@article{hou2026worldmodelsurvey,
  title={World Model for Robot Learning: A Comprehensive Survey},
  author={Hou, Bohan and Li, Gen and Jia, Jindou and An, Tuo and Guo, Xinying and Leng, Sicong and Geng, Haoran and Ze, Yanjie and Harada, Tatsuya and Torr, Philip and others},
  journal={arXiv preprint arXiv:2605.00080},
  year={2026}
}

@article{wang2026worldactionmodels,
  title={World Action Models: The Next Frontier in Embodied AI},
  author={Wang, Siyin and Shi, Junhao and Fu, Zhaoyang and He, Xinzhe and Liu, Feihong and Yang, Chenchen and Zhou, Yikang and Fei, Zhaoye and Gong, Jingjing and Fu, Jinlan and others},
  journal={arXiv preprint arXiv:2605.12090},
  year={2026}
}

@article{luo2026beingh07,
  title={Being-H0.7: A Latent World-Action Model from Egocentric Videos},
  author={Luo, Hao and Zhang, Wanpeng and Feng, Yicheng and Zheng, Sipeng and Xu, Haiweng and Xu, Chaoyi and Xi, Ziheng and Fu, Yuhui and Lu, Zongqing},
  journal={arXiv preprint arXiv:2605.00078},
  year={2026}
}

@article{liu2026oawam,
  title={OA-WAM: Object-Addressable World Action Model for Robust Robot Manipulation},
  author={Liu, Yushan and Sun, Peibo and Li, Shoujie and Xie, Yifan and Zhang, Lingfeng and Chao, Xintao and Dong, Shiyuan and Chen, Fang and Zhang, Xiao-Ping and Ding, Wenbo},
  journal={arXiv preprint arXiv:2605.06481},
  year={2026}
}

@article{zhao2023act,
  title={Learning fine-grained bimanual manipulation with low-cost hardware},
  author={Zhao, Tony Z and Kumar, Vikash and Levine, Sergey and Finn, Chelsea},
  journal={arXiv preprint arXiv:2304.13705},
  year={2023}
}

@inproceedings{zeng2021transporter,
  title={Transporter networks: Rearranging the visual world for robotic manipulation},
  author={Zeng, Andy and Florence, Pete and Tompson, Jonathan and Welker, Stefan and Chien, Jonathan and Attarian, Maria and Armstrong, Travis and Krasin, Ivan and Duong, Dan and Sindhwani, Vikas and others},
  booktitle={Conference on Robot Learning},
  pages={726--747},
  year={2021},
  organization={PMLR}
}

@inproceedings{shridhar2022cliport,
  title={Cliport: What and where pathways for robotic manipulation},
  author={Shridhar, Mohit and Manuelli, Lucas and Fox, Dieter},
  booktitle={Conference on robot learning},
  pages={894--906},
  year={2022},
  organization={PMLR}
}

@inproceedings{mo2021where2act,
  title={Where2act: From pixels to actions for articulated 3d objects},
  author={Mo, Kaichun and Guibas, Leonidas J and Mukadam, Mustafa and Gupta, Abhinav and Tulsiani, Shubham},
  booktitle={Proceedings of the IEEE/CVF International Conference on Computer Vision},
  pages={6813--6823},
  year={2021}
}

@inproceedings{wu2024gr1,
  title={Unleashing large-scale video generative pre-training for visual robot manipulation},
  author={Wu, Hongtao and Jing, Ya and Cheang, Chilam and Chen, Guangzeng and Xu, Jiafeng and Li, Xinghang and Liu, Minghuan and Li, Hang and Kong, Tao},
  booktitle={International Conference on Learning Representations},
  volume={2024},
  pages={10641--10662},
  year={2024}
}

@article{fei2025liberoplus,
  title={Libero-plus: In-depth robustness analysis of vision-language-action models},
  author={Fei, Senyu and Wang, Siyin and Shi, Junhao and Dai, Zihao and Cai, Jikun and Qian, Pengfang and Ji, Li and He, Xinzhe and Zhang, Shiduo and Fei, Zhaoye and others},
  journal={arXiv preprint arXiv:2510.13626},
  year={2025}
}

@article{zheng2025xvla,
  title={X-vla: Soft-prompted transformer as scalable cross-embodiment vision-language-action model},
  author={Zheng, Jinliang and Li, Jianxiong and Wang, Zhihao and Liu, Dongxiu and Kang, Xirui and Feng, Yuchun and Zheng, Yinan and Zou, Jiayin and Chen, Yilun and Zeng, Jia and others},
  journal={International Conference on Learning Representation},
  year={2026}
}

@article{brohan2022rt1,
  title={RT-1: Robotics Transformer for Real-World Control at Scale},
  author={Brohan, Anthony and Brown, Noah and Carbajal, Justice and Chebotar, Yevgen and Dabis, Joseph and Finn, Chelsea and Gopalakrishnan, Keerthana and Hausman, Karol and Herzog, Alexander and Hsu, Jasmine and others},
  journal={Robotics: Science and Systems XIX},
  year={2023},
  publisher={Robotics: Science and Systems Foundation}
}

@inproceedings{shridhar2023peract,
  title={Perceiver-actor: A multi-task transformer for robotic manipulation},
  author={Shridhar, Mohit and Manuelli, Lucas and Fox, Dieter},
  booktitle={Conference on Robot Learning},
  pages={785--799},
  year={2023},
  organization={PMLR}
}

@inproceedings{goyal2023rvt,
  title={Rvt: Robotic view transformer for 3d object manipulation},
  author={Goyal, Ankit and Xu, Jie and Guo, Yijie and Blukis, Valts and Chao, Yu-Wei and Fox, Dieter},
  booktitle={Conference on Robot Learning},
  pages={694--710},
  year={2023},
  organization={PMLR}
}

@InProceedings{gervet2023act3d,
  title = 	 {Act3D: 3D Feature Field Transformers for Multi-Task Robotic Manipulation},
  author =       {Gervet, Theophile and Xian, Zhou and Gkanatsios, Nikolaos and Fragkiadaki, Katerina},
  booktitle = 	 {Proceedings of The 7th Conference on Robot Learning},
  pages = 	 {3949--3965},
  year = 	 {2023},
  editor = 	 {Tan, Jie and Toussaint, Marc and Darvish, Kourosh},
  volume = 	 {229},
  series = 	 {Proceedings of Machine Learning Research},
  month = 	 {06--09 Nov},
  publisher =    {PMLR},
  url = 	 {https://proceedings.mlr.press/v229/gervet23a.html}
}

@article{ha2018world,
  title={World models},
  author={Ha, David and Schmidhuber, J{\"u}rgen},
  journal={arXiv preprint arXiv:1803.10122},
  year={2018}
}

@inproceedings{hafner2019planet,
  title={Learning latent dynamics for planning from pixels},
  author={Hafner, Danijar and Lillicrap, Timothy and Fischer, Ian and Villegas, Ruben and Ha, David and Lee, Honglak and Davidson, James},
  booktitle={International conference on machine learning},
  pages={2555--2565},
  year={2019},
  organization={PMLR}
}

@inproceedings{hafner2020dreamer,
  title={Dream to Control: Learning Behaviors by Latent Imagination},
  author={Hafner, Danijar and Lillicrap, Timothy and Ba, Jimmy and Norouzi, Mohammad},
  booktitle={International Conference on Learning Representations},
  year={2020}
}

@article{hafner2023dreamerv3,
  title={Mastering diverse domains through world models},
  author={Hafner, Danijar and Pasukonis, Jurgis and Ba, Jimmy and Lillicrap, Timothy},
  journal={arXiv preprint arXiv:2301.04104},
  year={2023}
}

@article{schrittwieser2020muzero,
  title={Mastering atari, go, chess and shogi by planning with a learned model},
  author={Schrittwieser, Julian and Antonoglou, Ioannis and Hubert, Thomas and Simonyan, Karen and Sifre, Laurent and Schmitt, Simon and Guez, Arthur and Lockhart, Edward and Hassabis, Demis and Graepel, Thore and others},
  journal={Nature},
  volume={588},
  number={7839},
  pages={604--609},
  year={2020},
  publisher={Nature Publishing Group UK London}
}

@inproceedings{finn2017visualforesight,
  title={Deep visual foresight for planning robot motion},
  author={Finn, Chelsea and Levine, Sergey},
  booktitle={2017 IEEE international conference on robotics and automation (ICRA)},
  pages={2786--2793},
  year={2017},
  organization={IEEE}
}

@article{ebert2018visualforesight,
  title={Visual foresight: Model-based deep reinforcement learning for vision-based robotic control},
  author={Ebert, Frederik and Finn, Chelsea and Dasari, Sudeep and Xie, Annie and Lee, Alex and Levine, Sergey},
  journal={arXiv preprint arXiv:1812.00568},
  year={2018}
}

@article{babaeizadeh2021fitvid,
  title={Fitvid: Overfitting in pixel-level video prediction},
  author={Babaeizadeh, Mohammad and Saffar, Mohammad Taghi and Nair, Suraj and Levine, Sergey and Finn, Chelsea and Erhan, Dumitru},
  journal={arXiv preprint arXiv:2106.13195},
  year={2021}
}

@inproceedings{hendrycks2019imagenetc,
  title={Benchmarking Neural Network Robustness to Common Corruptions and Perturbations},
  author={Hendrycks, Dan and Dietterich, Thomas},
  booktitle={International Conference on Learning Representations},
  year={2019}
}

@inproceedings{miller2021manyfaces,
  title={The Many Faces of Robustness: A Critical Analysis of Out-of-Distribution Generalization},
  author={Hendrycks, Dan and Basart, Steven and Mu, Norman and Kadavath, Saurav and Wang, Frank and Dorundo, Evan and Desai, Rahul and Zhu, Tyler and Parajuli, Samyak and Guo, Mike and others},
  booktitle={2021 IEEE/CVF International Conference on Computer Vision (ICCV)},
  pages={8320--8329},
  year={2021},
  organization={IEEE}
}

@inproceedings{nair2022r3m,
  title={R3M: A Universal Visual Representation for Robot Manipulation},
  author={Nair, Suraj and Rajeswaran, Aravind and Kumar, Vikash and Finn, Chelsea and Gupta, Abhinav},
  booktitle={Conference on Robot Learning},
  pages={892--909},
  year={2023},
  organization={PMLR}
}

@inproceedings{majumdar2023vip,
  title={VIP: Towards Universal Visual Reward and Representation via Value-Implicit Pre-Training},
  author={Ma, Yecheng Jason and Sodhani, Shagun and Jayaraman, Dinesh and Bastani, Osbert and Kumar, Vikash and Zhang, Amy},
  booktitle={The Eleventh International Conference on Learning Representations},
  year={2023}
}

@article{shah2023voltron,
  title={Language-driven representation learning for robotics},
  author={Karamcheti, Siddharth and Nair, Suraj and Chen, Annie S and Kollar, Thomas and Finn, Chelsea and Sadigh, Dorsa and Liang, Percy},
  journal={Robotics: Science and Systems},
  year={2023}
}

@inproceedings{black2025pi05,
  title={Pi0.5: a Vision-Language-Action Model with Open-World Generalization},
  author={Black, Kevin and Brown, Noah and Darpinian, James and Dhabalia, Karan and Driess, Danny and Esmail, Adnan and Equi, Michael Robert and Finn, Chelsea and Fusai, Niccolo and Galliker, Manuel Y and others},
  booktitle={9th Annual Conference on Robot Learning},
  year={2025}
}

@article{zhou2026gem4d,
  title = {GEM-4D: Geometry-Enhanced Video World Models for Robot Manipulation},
  author = {Zhou, Kaichen and Chen, Yuzhen and Zhan, Fangneng and Hua, Hang and Chen, Grace and Chang, Xinhai and Qu, Ao and Du, Yilun and Liu, Zhuang and Liang, Paul Pu and Wang, Mengyu},
  journal={arXiv preprint arXiv:2605.22882},
  year={2026}
}

@inproceedings{liu2025rdt,
  title={Rdt-1b: a diffusion foundation model for bimanual manipulation},
  author={Liu, Songming and Wu, Lingxuan and Li, Bangguo and Tan, Hengkai and Chen, Huayu and Wang, Zhengyi and Xu, Ke and Su, Hang and Zhu, Jun},
  booktitle={International Conference on Learning Representations},
  volume={2025},
  pages={29982--30009},
  year={2025}
}

@article{cen2025worldvla,
  title={Worldvla: Towards autoregressive action world model},
  author={Cen, Jun and Yu, Chaohui and Yuan, Hangjie and Jiang, Yuming and Huang, Siteng and Guo, Jiayan and Li, Xin and Song, Yibing and Luo, Hao and Wang, Fan and others},
  journal={arXiv preprint arXiv:2506.21539},
  year={2025}
}

@article{kim2025fine,
  title={Fine-tuning vision-language-action models: Optimizing speed and success},
  author={Kim, Moo Jin and Finn, Chelsea and Liang, Percy},
  journal={arXiv preprint arXiv:2502.19645},
  year={2025}
}

@article{yuan2026fastwam,
  title={Fast-wam: Do world action models need test-time future imagination?},
  author={Yuan, Tianyuan and Dong, Zibin and Liu, Yicheng and Zhao, Hang},
  journal={arXiv preprint arXiv:2603.16666},
  year={2026}
}

@article{zhang2026imagewam,
  title={ImageWAM: Do World Action Models Really Need Video Generation, or Just Image Editing?},
  author={Zhang, Yuyang and Zhang, Wenyao and Qi, Zekun and Zhang, He and Lin, Haitao and Zhang, Jingbo and Mu, Yao and Yang, Xiaokang and Zeng, Wenjun and Jin, Xin},
  journal={arXiv preprint arXiv:2606.19531},
  year={2026}
}

@article{hung2025nora,
  title={Nora: A small open-sourced generalist vision language action model for embodied tasks},
  author={Hung, Chia-Yu and Sun, Qi and Hong, Pengfei and Zadeh, Amir and Li, Chuan and Tan, U and Majumder, Navonil and Poria, Soujanya and others},
  journal={arXiv preprint arXiv:2504.19854},
  year={2025}
}

@article{cai2026internvla1,
  title={Internvla-a1: Unifying understanding, generation and action for robotic manipulation},
  author={Cai, Junhao and Cai, Zetao and Cao, Jiafei and Chen, Yilun and He, Zeyu and Jiang, Lei and Li, Hang and Li, Hengjie and Li, Yang and Liu, Yufei and others},
  journal={arXiv preprint arXiv:2601.02456},
  year={2026}
}

@article{ma2026internvla15,
  title={InternVLA-A1. 5: Unifying Understanding, Latent Foresight, and Action for Compositional Generalization},
  author={Ma, Haoxiang and Cai, Junhao and Xu, Xiaoxu and Li, Hao and Yang, Yuyin and Tian, Yang and Cao, Jiafei and Zhu, Hongrui and Qiu, Zherui and Yang, Yuqiang and others},
  journal={arXiv preprint arXiv:2607.04988},
  year={2026}
}

@article{bjorck2025gr00t,
  title={{GR00T N1}: An open foundation model for generalist humanoid robots},
  author={Bjorck, Johan and Casta{\~n}eda, Fernando and Cherniadev, Nikita and Da, Xingye and Ding, Runyu and Fan, Linxi and Fang, Yu and Fox, Dieter and Hu, Fengyuan and Huang, Spencer and others},
  journal={arXiv preprint arXiv:2503.14734},
  year={2025}
}

@article{li2024cogact,
  title={Cogact: A foundational vision-language-action model for synergizing cognition and action in robotic manipulation},
  author={Li, Qixiu and Liang, Yaobo and Wang, Zeyu and Luo, Lin and Chen, Xi and Liao, Mozheng and Wei, Fangyun and Deng, Yu and Xu, Sicheng and Zhang, Yizhong and others},
  journal={arXiv preprint arXiv:2411.19650},
  year={2024}
}

@article{team2025gemini,
  title={Gemini robotics: Bringing ai into the physical world},
  author={Team, Gemini Robotics and Abeyruwan, Saminda and Ainslie, Joshua and Alayrac, Jean-Baptiste and Arenas, Montserrat Gonzalez and Armstrong, Travis and Balakrishna, Ashwin and Baruch, Robert and Bauza, Maria and Blokzijl, Michiel and others},
  journal={arXiv preprint arXiv:2503.20020},
  year={2025}
}

@article{black2026real,
  title={Real-time execution of action chunking flow policies},
  author={Black, Kevin and Galliker, Manuel and Levine, Sergey},
  journal={Advances in Neural Information Processing Systems},
  volume={38},
  pages={33383--33407},
  year={2026}
}

@article{wang2026qwenvla,
  title={Qwen-vla: Unifying vision-language-action modeling across tasks, environments, and robot embodiments},
  author={Wang, Qiuyue and Li, Mingsheng and Guan, Jian and Ye, Jinhui and Xie, Sicheng and Liu, Yitao and Chen, Junhao and Liang, Zhixuan and Zhang, Jie and Hu, Xintong and others},
  journal={arXiv preprint arXiv:2605.30280},
  year={2026}
}

@article{yuan2026qwenrobotmanip,
  title={Qwen-robotmanip technical report: Alignment unlocks scale for robotic manipulation foundation models},
  author={Yuan, Haoqi and Liang, Zhixuan and Chen, Anzhe and Wang, Ye and Li, Haoyang and Lin, Pei and Huang, Yiyang and Lei, Zixing and Zhang, Tong and Zhang, Jiazhao and others},
  journal={arXiv preprint arXiv:2606.17846},
  year={2026}
}

@article{cheang2025gr3,
  title={Gr-3 technical report},
  author={Cheang, Chilam and Chen, Sijin and Cui, Zhongren and Hu, Yingdong and Huang, Liqun and Kong, Tao and Li, Hang and Li, Yifeng and Liu, Yuxiao and Ma, Xiao and others},
  journal={arXiv preprint arXiv:2507.15493},
  year={2025}
}

@article{bu2025univla,
  title={Univla: Learning to act anywhere with task-centric latent actions},
  author={Bu, Qingwen and Yang, Yanting and Cai, Jisong and Gao, Shenyuan and Ren, Guanghui and Yao, Maoqing and Luo, Ping and Li, Hongyang},
  journal={arXiv preprint arXiv:2505.06111},
  year={2025}
}

@article{wu2026linbotvla,
  title={A pragmatic vla foundation model},
  author={Wu, Wei and Lu, Fan and Wang, Yunnan and Yang, Shuai and Liu, Shi and Wang, Fangjing and Zhu, Qian and Sun, He and Wang, Yong and Ma, Shuailei and others},
  journal={arXiv preprint arXiv:2601.18692},
  year={2026}
}

@inproceedings{bi2026motus,
  title={Motus: A unified latent action world model},
  author={Bi, Hongzhe and Tan, Hengkai and Xie, Shenghao and Wang, Zeyuan and Huang, Shuhe and Liu, Haitian and Zhao, Ruowen and Feng, Yao and Xiang, Chendong and Rong, Yinze and others},
  booktitle={Proceedings of the IEEE/CVF Conference on Computer Vision and Pattern Recognition},
  pages={35101--35113},
  year={2026}
}

@article{li2026lingbot-va,
  title={Causal world modeling for robot control},
  author={Li, Lin and Zhang, Qihang and Luo, Yiming and Yang, Shuai and Wang, Ruilin and Han, Fei and Yu, Mingrui and Gao, Zelin and Xue, Nan and Zhu, Xing and others},
  journal={arXiv preprint arXiv:2601.21998},
  year={2026}
}

@article{kim2026cosmos,
  title={Cosmos policy: Fine-tuning video models for visuomotor control and planning},
  author={Kim, Moo Jin and Gao, Yihuai and Lin, Tsung-Yi and Lin, Yen-Chen and Ge, Yunhao and Lam, Grace and Liang, Percy and Song, Shuran and Liu, Ming-Yu and Finn, Chelsea and others},
  journal={arXiv preprint arXiv:2601.16163},
  year={2026}
}

@article{zhao2026faster1,
  title={Faster-WAM: Efficient Inference-Time Future Conditioning for Robust World Action Models},
  author={Zhao, Weiheng and Jiang, Haoyi and Shi, Xin and Liu, Liu and Huang, Fan and Su, Zhizhong and Sui, Wei and Wang, Xinggang},
  journal={arXiv preprint arXiv:2608.04404},
  year={2026}
}

@article{ma2026faster2,
  title={Faster-WAM: Do World Action Models Need Deep Action Modules?},
  author={Ma, Liheng and Yang, Rui Heng and Zhang, Zhanguang and Clemente, Mateo and Hu, Ziwen and Cao, Tongtong and Zhang, Yingxue},
  journal={arXiv preprint arXiv:2608.02365},
  year={2026}
}

@article{wan2025wan,
  title={Wan: Open and advanced large-scale video generative models},
  author={Wan, Team and Wang, Ang and Ai, Baole and Wen, Bin and Mao, Chaojie and Xie, Chen-Wei and Chen, Di and Yu, Feiwu and Zhao, Haiming and Yang, Jianxiao and others},
  journal={arXiv preprint arXiv:2503.20314},
  year={2025}
}

@article{simeoni2025dinov3,
  title={Dinov3},
  author={Sim{\'e}oni, Oriane and Vo, Huy V and Seitzer, Maximilian and Baldassarre, Federico and Oquab, Maxime and Jose, Cijo and Khalidov, Vasil and Szafraniec, Marc and Yi, Seungeun and Ramamonjisoa, Micha{\"e}l and others},
  journal={arXiv preprint arXiv:2508.10104},
  year={2025}
}

@article{ye2026dreamzero,
  title={World action models are zero-shot policies},
  author={Ye, Seonghyeon and Ge, Yunhao and Zheng, Kaiyuan and Gao, Shenyuan and Yu, Sihyun and Kurian, George and Indupuru, Suneel and Tan, You Liang and Zhu, Chuning and Xiang, Jiannan and others},
  journal={arXiv preprint arXiv:2602.15922},
  year={2026}
}

@inproceedings{perez2018film,
  title={Film: Visual reasoning with a general conditioning layer},
  author={Perez, Ethan and Strub, Florian and De Vries, Harm and Dumoulin, Vincent and Courville, Aaron},
  booktitle={Proceedings of the AAAI conference on artificial intelligence},
  volume={32},
  number={1},
  year={2018}
}

@inproceedings{lipman2023flow,
  title={Flow Matching for Generative Modeling},
  author={Lipman, Yaron and Chen, Ricky TQ and Ben-Hamu, Heli and Nickel, Maximilian and Le, Matt},
  booktitle={International Conference on Learning Representations},
  year={2023},
  url={https://arxiv.org/abs/2210.02747}
}

@inproceedings{farneback2003two,
  title={Two-frame motion estimation based on polynomial expansion},
  author={Farneb{\"a}ck, Gunnar},
  booktitle={Scandinavian conference on Image analysis},
  pages={363--370},
  year={2003},
  organization={Springer}
}

@article{zhao2025statefree,
  title={Do you need proprioceptive states in visuomotor policies?},
  author={Zhao, Juntu and Lu, Wenbo and Zhang, Di and Liu, Yufeng and Liang, Yushen and Zhang, Tianluo and Cao, Yifeng and Xie, Junyuan and Hu, Yingdong and Wang, Shengjie and others},
  journal={arXiv preprint arXiv:2509.18644},
  year={2025}
}

@article{lu2026would,
  title={When would vision-proprioception policies fail in robotic manipulation?},
  author={Lu, Jingxian and Xia, Wenke and Wu, Yuxuan and Lu, Zhiwu and Hu, Di},
  journal={arXiv preprint arXiv:2602.12032},
  year={2026}
}

@inproceedings{xuhong2018explicit,
  title={Explicit inductive bias for transfer learning with convolutional networks},
  author={Xuhong, LI and Grandvalet, Yves and Davoine, Franck},
  booktitle={International conference on machine learning},
  pages={2825--2834},
  year={2018},
  organization={PMLR}
}

@inproceedings{xiao2024florence,
  title={Florence-2: Advancing a unified representation for a variety of vision tasks},
  author={Xiao, Bin and Wu, Haiping and Xu, Weijian and Dai, Xiyang and Hu, Houdong and Lu, Yumao and Zeng, Michael and Liu, Ce and Yuan, Lu},
  booktitle={2024 IEEE/CVF Conference on Computer Vision and Pattern Recognition (CVPR)},
  pages={4818--4829},
  year={2024},
  organization={IEEE}
}

@misc{qwen35blog,
    title = {Qwen3.5: Towards Native Multimodal Agents},
    url = {https://qwen.ai/blog?id=qwen3.5},
    author = {Qwen Team},
    month = {February},
    year = {2026}
}

@article{zhao2026sgwam,
  title={SG-WAM: Self-Guided World Modeling in Geometry-Aware Policy Space},
  author={Zhao, Ruiteng and Zhang, Zhengshen and Su, Yue and Wang, Wenshuo and Li, Jiahui and Yang, Zhiyuan and Tay, Francis EH and Ang Jr, Marcelo H and Zhu, Haiyue},
  journal={arXiv preprint arXiv:2608.01397},
  year={2026}
}

@article{zeng2026enfold,
  title={Enfold: Folding World Model Imagination into Predictive Representations for Ultra-Efficient Embodied Control},
  author={Zeng, Weili and Xing, Yitong and Liu, Fulong and Yang, Chengqun and Xiang, Antao and Tian, Feng and Gao, Jingnan and Cai, Jisong and Wang, Xin and Wu, Xiaomin and others},
  journal={arXiv preprint arXiv:2607.26657},
  year={2026}
}

@article{beyer2024paligemma,
  title={Paligemma: A versatile 3b vlm for transfer},
  author={Beyer, Lucas and Steiner, Andreas and Pinto, Andr{\'e} Susano and Kolesnikov, Alexander and Wang, Xiao and Salz, Daniel and Neumann, Maxim and Alabdulmohsin, Ibrahim and Tschannen, Michael and Bugliarello, Emanuele and others},
  journal={arXiv preprint arXiv:2407.07726},
  year={2024}
}

@article{community2026starvla,
  title={StarVLA: A Lego-like Codebase for Vision-Language-Action Model Developing},
  author={Community, StarVLA},
  journal={arXiv preprint arXiv:2604.05014},
  year={2026}
}

@article{yang2026abot,
  title={Abot-m0: Vla foundation model for robotic manipulation with action manifold learning},
  author={Yang, Yandan and Zeng, Shuang and Lin, Tong and Chang, Xinyuan and Qi, Dekang and Xiao, Junjin and Liu, Haoyun and Chen, Ronghan and Chen, Yuzhi and Huo, Dongjie and others},
  journal={arXiv preprint arXiv:2602.11236},
  year={2026}
}

@article{lin2026holobrain,
  title={Holobrain-0 technical report},
  author={Lin, Xuewu and Lin, Tianwei and Du, Yun and Xie, Hongyu and Jin, Yiwei and Li, Jiawei and Wu, Shijie and Wang, Qingze and Li, Mengdi and Zhao, Mengao and others},
  journal={arXiv preprint arXiv:2602.12062},
  year={2026}
}

@inproceedings{xiao2026ava,
  title={AVA-VLA: improving vision-language-action models with active visual attention},
  author={Xiao, Lei and Li, Jifeng and Gao, Juntao and Ye, Feiyang and Jin, Yan and Qian, Jingjing and Zhang, Jing and Wu, Yong and Yu, Xiaoyuan},
  booktitle={Proceedings of the IEEE/CVF Conference on Computer Vision and Pattern Recognition},
  pages={13453--13463},
  year={2026}
}

@article{tan2025RIPT-VLA,
  title={Interactive post-training for vision-language-action models},
  author={Tan, Shuhan and Dou, Kairan and Zhao, Yue and Kr{\"a}henb{\"u}hl, Philipp},
  journal={arXiv preprint arXiv:2505.17016},
  year={2025}
}

@inproceedings{liao2026geact,
  title={Genie envisioner: A unified world foundation platform for robotic manipulation},
  author={Liao, Yue and Zhou, Pengfei and Huang, Siyuan and Yang, Donglin and Chen, Shengcong and Jiang, Yuxin and Hu, Yue and Liu, Si and Luo, Jianlan and Chen, Liliang and others},
  booktitle={International Conference on Learning Representations},
  volume={2026},
  pages={88446--88463},
  year={2026}
}

@article{pertsch2025pi0fast,
  title={Fast: Efficient action tokenization for vision-language-action models},
  author={Pertsch, Karl and Stachowicz, Kyle and Ichter, Brian and Driess, Danny and Nair, Suraj and Vuong, Quan and Mees, Oier and Finn, Chelsea and Levine, Sergey},
  booktitle = {Robotics: Science and Systems},
  year      = {2025}
}

@inproceedings{sun2026vlajepa,
  title={Vla-jepa: Enhancing vision-language-action model with latent world model},
  author={Sun, Jingwen and Zhang, Wenyao and Qi, Zekun and Ren, Shaojie and Liu, Zezhi and Zhu, Hanxin and Sun, Guangzhong and Jin, Xin and Chen, Zhibo},
  booktitle={European Conference on Computer Vision},
  pages={478--497},
  year={2026},
  organization={Springer}
}
\bibliographystyle{iclr2027_conference}

\appendix
\clearpage
\appendix

\begin{center}
    {\Large\bfseries Appendix}
\end{center}

\section{Notation and Symbol Definitions}
\label{app:notation}
Table~\ref{tab:notation} lists additional variables introduced in the appendices.

\begingroup
\small
\setlength{\LTleft}{0pt}
\setlength{\LTright}{0pt}
\begin{longtable}{@{}p{0.29\linewidth}p{0.67\linewidth}@{}}
\caption{Additional notation used in the appendices.}\label{tab:notation}\\
\toprule Symbol & Meaning \\
\midrule
\endfirsthead
\multicolumn{2}{l}{Table~\ref{tab:notation} (continued)}\\
\toprule Symbol & Meaning \\
\midrule
\endhead
\midrule\multicolumn{2}{r}{Continued on next page}\\
\endfoot
\bottomrule
\endlastfoot
\multicolumn{2}{l}{\textbf{World targets and distillation}}\\
$h,h'$ & Future-frame offset and alternative offset in a horizon sum. \\
$k;\ FW,FD,CW,CD,M$ & Prediction-head index; WAN future, DINO future, WAN change, DINO change, and motion flow, respectively. \\
$W,D$ in $Z^W,Z^D$ & WAN-VAE and DINO feature sources, respectively. \\
$Z^W_t,Z^D_t$ & Encoded current WAN and DINO features. \\
$Z^W_{t,h},Z^D_{t,h}$ & Last WAN temporal latent of a clip ending at $t+h$, and DINO features of that future frame. \\
$Y^k_{t,h},\bar Y^k_t,\widehat Y^k_t$ & Target at one horizon, horizon-fused target, and current-context prediction for field $k$. \\
$m_{t,h}$ & Binary indicator of an available within-episode future. Distinct from memory-type index $m$. \\
$\xi_h^k,\pi_{t,h}^k$ & Learned horizon logit and its normalized, validity-masked mixture weight. \\
$\pi^{k,0},\pi^k$ & Initial and learned horizon distributions used by the mixture regularizer; distinct from policy $\pi_\theta$. \\
$P_t,Q_k,h_k$ & Valid current multimodal prefix, learned queries, and output mapping of prediction head $k$. The mapping $h_k$ is distinct from horizon $h$. \\
$\lambda_2^k,\lambda_{\cos}^k$ & Nonnegative coefficients for field-wise mean squared error and cosine distance. \\
$\mathcal L_{\mathrm{bal}},\lambda_{\mathrm{bal}}$ & Horizon-prior KL regularizer and its coefficient. \\
\multicolumn{2}{l}{\textbf{Memory formation and attention}}\\
$\mathcal S_m,U_k$ & Set of sources fused into memory $m$, and representation of source $k$. Here $k$ ranges over fusion sources, including foresight, not just prediction heads. \\
$W_k,e_k,N_m$ & Source projection, source identity embedding, and output token count for memory $m$. \\
$\zeta^m,\alpha_k^m$ & Source-fusion logits and normalized mixture weight of source $k$. \\
$\mathcal C_t^x,\phi_x$ & Token source indexed by $x\in\{P,F,C,M\}$ and its projection into the shared global-summary space. \\
$\boldsymbol\zeta^g,\alpha_x^g$ & Trainable global-summary fusion logits and normalized weight of source $x$. \\
\multicolumn{2}{l}{\textbf{Reliability, residuals, and state guidance}}\\
$\bar H_l,\bar C_a^m$ & Pooled action features and pooled memory features projected to a compatible action-feature space. Subscript $a$ here means action-aligned, not augmented. \\
$f_m,b_m^r,r_l^m$ & Learned compatibility mapping, scalar reliability bias, and attention reliability gate for memory $m$ at layer $l$. \\
$u_m,u_{l,m},c_m$ & Modality gate logit, layer-specific modality gate logit, and channel-wise gate vector. \\
$\mathcal I_m$ & Set of transformer layers receiving memory $m$. \\
$\Delta_l^m$ & Gated attention residual for memory $m$ at layer $l$. \\
$q(s_t),q_1,\ldots,q_4$ & Bounded state-conditioned scalars that rescale future, change, motion, and global context, respectively. \\
$p_t$ & Demonstration-derived transition score used to scale gradients through the state-guidance path. \\
\multicolumn{2}{l}{\textbf{Robust learning and optimization}}\\
$v,\widetilde v$ & Clean and augmented predicted action velocities in the consistency loss; shorthand for evaluations of $v_\theta$. \\
$V,\bar V,\widetilde{\bar V}$ & Visual feature tensor, pooled clean features, and pooled augmented features in consistency/perturbation equations. Here $V$ is not the number of cameras. \\
$S_i,\widetilde S_i,K_s,i$ & Clean slot, augmented slot, number of slots, and slot index. \\
$\lambda_{\mathrm{con},a},\lambda_{\mathrm{con},v}$ & Action and visual consistency weights. \\
$\lambda_{\mathrm{con},w},\lambda_{\mathrm{con},s}$ & World-memory and slot consistency weights. \\
$\delta_V,G,\epsilon_V$ & Visual perturbation, action-loss gradient with respect to visual features, and relative perturbation magnitude. \\
$j,\eta_j,\lambda_j$ & Optimizer parameter-group index, learning rate, and source-anchor strength. \\
$\theta_j^{\mathrm{src}},\widetilde\theta_j,\theta_j^+$ & Fixed source parameters, parameters after the optimizer update but before anchoring, and parameters after anchoring. Here the tilde does not denote augmentation. \\
\end{longtable}
\endgroup

\section{World-Prior Distillation and Conditioning}
\label{app:technical}

\subsection{Demonstration-derived supervision}
Let $Z^W_{t,h}$ denote the last temporal WAN-VAE latent of a causal clip ending at $t+h$, and $Z^D_{t,h}$ the DINOv3 patch features of that future frame. With corresponding current features $Z^W_t,Z^D_t$, the five supervision fields are
\begin{align}
 Y^{FW}_{t,h}&=\Pool_W(Z^W_{t,h}), &
 Y^{FD}_{t,h}&=\Pool_D(Z^D_{t,h}),\nonumber\\
 Y^{CW}_{t,h}&=\Pool_2\!\left(\mathcal N(\|Z^W_{t,h}-Z^W_t\|_2)\right), &
 Y^{CD}_{t,h}&=\Pool_2\!\left(\mathcal N(\|Z^D_{t,h}-Z^D_t\|_2)\right),\nonumber\\
 Y^M_{t,h}&=\Pool_2\!\left(\operatorname{Flow}(I_t,I_{t+h})\right).&&
\end{align}
Here $\mathcal N$ normalizes each change map by its spatial maximum. $\Pool_2$ denotes spatial pooling, whereas future-token pooling operates on the feature sequence. Optical flow~\citep{farneback2003two} is normalized by image dimensions. These targets encode observed temporal changes rather than identified causal effects.

To aggregate complementary horizons, we learn a masked mixture for each field:
\begin{equation}
 \pi^k_{t,h}=\frac{m_{t,h}\exp\xi^k_h}{\sum_{h'}m_{t,h'}\exp\xi^k_{h'}},
 \qquad \bar Y^k_t=\sum_h\pi^k_{t,h}\sg(Y^k_{t,h}).
\end{equation}
The validity mask $m$ excludes futures outside the episode; samples without a valid future receive no distillation loss. The teacher encoders are frozen, while mixture weights are learned. With $d_{\cos}(x,y)=1-\cos(\operatorname{vec}x,\operatorname{vec}y)$, the objective is
\begin{equation}
 \mathcal L_W=\sum_k\left[\lambda^k_2\operatorname{MSE}(\widehat Y^k_t,\bar Y^k_t)
 +\lambda^k_{\cos}d_{\cos}(\widehat Y^k_t,\bar Y^k_t)\right]
 +\lambda_{\mathrm{bal}}\mathcal L_{\mathrm{bal}},
 \label{eq:worldloss}
\end{equation}
where $\mathcal L_{\mathrm{bal}}$ averages $D_{\mathrm{KL}}(\pi^{k,0}\|\pi^k)$ over fields and encourages retention of the initial horizon prior.

\subsection{Prediction and memory formation}
\label{app:memory}
For a valid current-context prefix $P_t$, excluding padding and teacher-forced actions, each predictor uses learned queries $Q_k$:
\begin{equation}
 \widehat Y^k_t=h_k\!\left(\operatorname{CrossAttn}(Q_k,\sg(P_t),\sg(P_t))\right),
 \quad k\in\{FW,FD,CW,CD,M\}.
\end{equation}
The operator $h_k$ contains residual processing and a field-specific projection; $\sg$ denotes stop-gradient. Projected source representations $U_k$ are normalized, pooled, and fused into memories:
\begin{equation}
 C^m=\sum_{k\in\mathcal S_m}\alpha^m_k
 [\LN(\Pool_{N_m}(U_kW_k))+e_k],\qquad
 \alpha^m=\softmax(\zeta^m),\quad m\in\{F,C\}.
\end{equation}
Future sources comprise predicted DINO features and the backbone's trainable foresight representation; change sources comprise WAN and DINO change predictions. Motion uses a projection of the predicted flow. Foresight tokens remain active without video-denoising supervision.

To provide global FiLM conditioning, we pool and project the current prefix and the three world memories, then combine them with learned source weights:
\begin{equation}
 g_t=\sum_{x\in\{P,F,C,M\}}\alpha_x^g
 \phi_x\!\left(\Pool(\mathcal C_t^x)\right),\qquad
 \boldsymbol\alpha^g=\softmax(\boldsymbol\zeta^g),
 \quad \mathcal C_t^P=P_t,\quad \mathcal C_t^m=C^m.
 \label{eq:global_world}
\end{equation}
Here $x$ indexes the current prefix ($P$) or a future, change, or motion source. $\mathcal C_t^x$ denotes the corresponding token sequence, $\phi_x$ is a source-specific projection into a shared feature space, and $\alpha_x^g$ is the normalized weight obtained from trainable logits $\boldsymbol\zeta^g$. Consequently, $g_t$ is a global summary computed from current-context predictions rather than an additional supervision target or cached future input.

\subsection{Selective world-to-action conditioning}
The action transformer allocates overlapping layer intervals to future, change, and motion guidance, progressing from outcome information to local refinement. Within a layer, active memories update the action stream sequentially. Pooled action and projected memory features define a compatibility score:
\begin{equation}
 r_l^m=\sigmoid\!\left(f_m[\bar H_l,\bar C_a^m,
 \bar C_a^m-\bar H_l,\bar H_l\odot\bar C_a^m]+b_m^r\right).
\end{equation}
With channel gate $c_m$, modality logit $u_m$, and layer logit $u_{l,m}$, the attention contribution is
\begin{equation}
 \Delta_l^m=\mathbf{1}_{\{l\in\mathcal I_m\}}\,
 \eta\sigmoid(u_m)\sigmoid(u_{l,m})r_l^m(c_m\odot R_l^m).
\end{equation}
Reliability is learned compatibility, not calibrated uncertainty. Change biases use predicted WAN change scores, while motion biases use displacement magnitudes. Applying Eq.~\eqref{eq:attn_residual} to these gated contributions gives $\Delta_l^{\mathrm{attn}}$. FiLM has a separate reliability gate computed from action and modulation features. Its affine transformation acts on the original layer input; the adapter processes $\Delta_l=\Delta_l^{\mathrm{attn}}+\Delta_l^{\mathrm{film}}$. Residual output projections are zero-initialized.

\paragraph{Current evidence and state guidance}
Language-conditioned visual slots complement predicted dynamics. Each slot contains a learned address and visual content retrieved with camera-aware queries. Actions query slot addresses as keys and full slots as values. In the LIBERO setting, slots and world targets use the main camera, while the VLM retains both main and wrist observations. A trainable visual adapter supports adaptation of the frozen visual pathway.

Normalized robot state modulates guidance through bounded scalars:
\begin{equation}
 q(s_t)=0.1\tanh(\operatorname{MLP}(\LN(s_t)))\in\R^4.
\end{equation}
These rescale $(C^F,C^C,C^M,g_t)$ by $(1+q_1,\ldots,1+q_4)$. To limit reliance on proprioceptive shortcuts~\citep{zhao2025statefree,lu2026would}, a demonstration-derived transition score $p_t$ scales gradients through this guidance path by $1-0.5p_t$, without changing its forward computation or requiring transition labels at inference.

\section{Robust Learning and Experimental Protocol}
\label{app:learning}

\subsection{Paired consistency and regularization}
Clean and augmented observations share action labels, flow noise, and interpolation time. State perturbations are aligned across numeric and tokenized inputs, preserving gripper information. Task-preserving language rewrites and photometric perturbations alter observations without relocating objects or changing action labels. The consistency objective is
\begin{align}
 \mathcal L_{\mathrm{con}}={}&\lambda_{\mathrm{con},a}
 \operatorname{SmoothL1}(\widetilde v,\sg(v))
 +\lambda_{\mathrm{con},v}d_{\cos}(\widetilde{\bar V},\sg(\bar V))\nonumber\\
 &+\frac{\lambda_{\mathrm{con},w}}{3}\sum_m
 d_{\cos}(\Pool(\widetilde C^m),\sg(\Pool(C^m)))
 +\frac{\lambda_{\mathrm{con},s}}{K_s}\sum_i
 d_{\cos}(\widetilde S_i,\sg(S_i)).
\end{align}
Here tildes denote augmented features, $\bar V$ denotes pooled visual features, and $S_i$ denotes a slot. Clean targets are detached for consistency only; clean action and world supervision remain differentiable. To regularize feature sensitivity, we additionally perturb augmented visual features along a detached action-loss gradient direction:
\begin{equation}
 \delta_V=\epsilon_V\|V\|_2\sg\!\left(G/\max(\|G\|_2,\varepsilon)\right),
 \qquad G=\nabla_V\mathcal L_{\mathrm{act}}.
\end{equation}
Norms are computed per example. Demonstration-derived start, state-rarity, and motion-transition scores mildly reweight supervision toward underrepresented or changing states. Such reweighting preserves demonstrated action labels rather than synthesizing recovery trajectories.

To retain pretrained behavior during adaptation, we use a proximal parameter anchor~\citep{xuhong2018explicit}. Following an optimizer update to $\widetilde\theta_j$, parameter group $j$ is updated as
\begin{equation}
 \theta_j^+=\frac{\widetilde\theta_j+\eta_j\lambda_j\theta_j^{\mathrm{src}}}
 {1+\eta_j\lambda_j},
\end{equation}
where $\theta^{\mathrm{src}}$ is the fixed initialization and $\eta_j$ is the learning rate. This regularizes parameter drift without an additional teacher-policy forward pass.

\subsection{Training Configuration}
World targets use horizons of 16, 24, and 30 frames; these horizons describe scene evolution beyond the execution chunk. The predictor uses 16 queries per field, and world attention operates at width 256. In the 24-layer action transformer, zero-indexed future/change/motion intervals are $[0,16)$, $[8,24)$, and $[16,24)$.



\section{Simulation Benchmarks and Real-World Setup}
\label{app:sim_real_setup}

\subsection{Simulation Benchmarks}
\label{app:benchmark_protocols}

We evaluate \method{} on LIBERO-Plus, our LIBERO-Dynamic, RoboTwin~2.0, and VLABench, as shown in Fig.~\ref{fig:exp_setup_vis}(c). We train exclusively on the official LIBERO demonstrations~\citep{liu2023libero}, using language instructions, third-person and wrist-camera RGB observations, and robot proprioception. Each action is represented as $\mathbf{a}_t=[\Delta\mathbf{p}_t,\Delta\mathbf{r}_t,g_t]\in\mathbb{R}^{7}$, where $\Delta\mathbf{p}_t,\Delta\mathbf{r}_t\in\mathbb{R}^{3}$ are end-effector translation and rotation commands and $g_t$ is the gripper command. The policy predicts an $T_a=8$ action chunk. \textbf{LIBERO-Plus}~\citep{fei2025liberoplus} evaluates the same LIBERO-only trained policy on seven controlled shifts: camera viewpoint, robot initial state, language instruction, lighting, background, visual noise, and scene layout. We report success rates for each LIBERO-Plus perturbation. Unlike the static shifts in LIBERO-Plus, \textbf{LIBERO-Dynamic} evaluates closed-loop robustness to time-varying camera viewpoints. It extends the 40 canonical LIBERO tasks with five predefined camera-motion profiles: \textit{smooth panning}, \textit{orbital motion}, \textit{continuous zoom}, \textit{abrupt viewpoint jumps}, and \textit{camera shake}. Only the third-person camera moves; the wrist camera and success criteria remain unchanged. We evaluate 50 initial states per task over five balanced difficulty levels, yielding 10,000 episodes with policy-independent camera viewpoint trajectories. No LIBERO-Dynamic data are used for training, and success rates are reported by camera-motion profile. \textbf{RoboTwin~2.0}~\citep{chen2025robotwin2} evaluates language-conditioned bimanual manipulation under \textit{Easy} and \textit{Hard} settings. Easy contains clean scenes, while Hard randomizes distractor objects, background textures, lighting, tabletop height, and language instructions. We train on the official 50 tasks using 50 clean and 500 randomized trajectories per task. Inputs comprise the instruction, head-camera and bilateral wrist-camera RGB observations, and proprioception. Actions are $\mathbf{a}_t=[\mathbf{q}^{L}_t,g^{L}_t,\mathbf{q}^{R}_t,g^{R}_t]\in\mathbb{R}^{14}$, where $\mathbf{q}^{L}_t,\mathbf{q}^{R}_t\in\mathbb{R}^{6}$ are left- and right-arm joint commands and $g^{L}_t,g^{R}_t$ are gripper commands. The action-chunk horizon is $T_a=50$. We evaluate 100 episodes per task with unseen instruction variants and report mean success rates separately for Easy and Hard. \textbf{VLABench}~\citep{zhang2024vlabench} evaluates single-arm manipulation under in-distribution and systematic generalization settings. We train on its official primitive-task dataset containing 10 tasks and 500 demonstrations per task. Inputs include the instruction, front-, secondary-, and wrist-camera RGB observations, and proprioception. Each action is $\mathbf{a}_t=[\mathbf{p}_t,\mathbf{r}^{6\mathrm{D}}_t,g_t]\in\mathbb{R}^{10}$, where $\mathbf{p}_t\in\mathbb{R}^{3}$ is the end-effector position and $\mathbf{r}^{6\mathrm{D}}_t\in\mathbb{R}^{6}$ is its continuous rotation representation; the action-chunk horizon is $T_a=30$. We evaluate policy on five tracks: \textit{In-Distribution}, \textit{Cross-Category}, \textit{Common Sense}, \textit{Semantic Instruction}, and \textit{Unseen Texture}. The latter four test zero-shot transfer to unseen categories, implicit or knowledge-dependent instructions, and appearance shifts. We report success rate (SR), intention score (IS), and progress score (PS), averaged across tracks.

\subsection{Real-Robot Experimental Setup, Task Definitions, and Evaluation Protocol}
\label{app:real_robot_setup}

\begin{figure}[h]
    \includegraphics[width=1\linewidth]{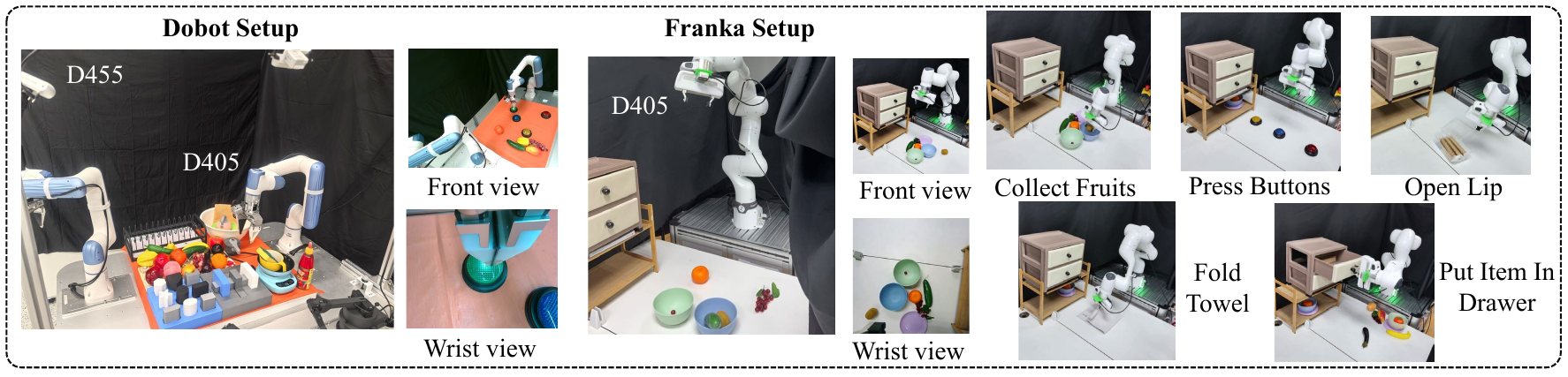}
    \vspace{-16pt}
    \caption{Dobot and Franka real-world setup.}
    \label{fig:dobot-franka}
\end{figure}

\paragraph{Overview}
We evaluate the real-robot task execution capability and OOD generalization capability of \method{} on two single-arm platforms: a Dobot Nova2 robot and a Franka FR3 robot. Both platforms use LeRobot-style data with synchronized third-person RGB, wrist RGB, Cartesian robot state, Cartesian end-effector action, timestamps, and task instructions. The action vector is seven-dimensional, $\mathbf{a}_t=[x_t,y_t,z_t,\mathrm{roll}_t,\mathrm{pitch}_t,\mathrm{yaw}_t,g_t]\in\mathbb{R}^{7}$, where $g_t\in[0,1]$ denotes the gripper command. Demonstrations are collected at 30 Hz. The Dobot setup uses an Intel RealSense D455 as the third-person camera and an Intel RealSense D405 as the wrist camera, with $480\times480$ videos for both views. The Franka setup uses an ORBBEC Femto Bolt as the third-person camera at $720\times720$ resolution and an Intel RealSense D405 as the wrist camera at $480\times480$ resolution. At training time, images are resized to the policy resolution and the same action-chunk interface is used across embodiments. The robot is reset to a fixed home pose before each rollout. Object poses are randomly sampled within task-specific regions, while preserving reachability and avoiding initial collisions. The dataset contains 50 demonstrations per task.

\subsubsection{Real-Robot Task Definitions}
\label{app:real_task_definitions}
\textbf{Pick Grape} On the table, there is a bunch of grapes, other fruits, and a plate. The robot needs to pick up and place the grapes. Language instruction: Put the grapes into the $\mathrm{[COLOR]}$ plate.

\textbf{Collect Fruits} There are various fruits on the table. The robot needs to collect the specified fruits into the specific target. Language instruction: Put a $\mathrm{[FRUIT]}$, a $\mathrm{[FRUIT]}$, and an $\mathrm{[FRUIT]}$ into the $\mathrm{[COLOR]}$ $\mathrm{[TARGET]}$.

\textbf{Push Buttons/Press Buttons} There are multiple buttons of different colors on the table. The robot needs to press the buttons in a specific sequence. Language instruction: Press the $\mathrm{[COLOR]}$ button, then the $\mathrm{[COLOR]}$ button, followed by the $\mathrm{[COLOR]}$ button, and finally the $\mathrm{[COLOR]}$ button.

\textbf{Stack Bowls} There are a cup, some bowls and a plate on the table. The robot needs to stack a cup, the bowls and plate in a specific order. Language instruction: Stack a $\mathrm{[COLOR]}$ cup,a $\mathrm{[COLOR]}$ bowl and a $\mathrm{[COLOR]}$ bowl on a $\mathrm{[COLOR]}$ plate.

\textbf{Put Item In Drawer} There is a cabinet on the table. The robot needs to open the specified drawer, place the target object inside, and then close the drawer. Language instruction: Open the $\mathrm{[Top/Middle/Bottom]}$ drawer and put the $\mathrm{[OBJECT]}$ into it.

\textbf{Fold Towel} There is a towel on the table. The robot needs to pick up one corner of the towel and then fold it. Language instruction: Fold the towel.

\textbf{Open Lip} There is a box on the table. The robot needs to pick up the lid and open it. Language instruction: Open the plastic lid of the biscuit case. 

\paragraph{Hard-track perturbations}
For Hard-track evaluation, we introduce controlled visual and scene perturbations while keeping the task goal and success criterion unchanged. In the distractor setting, additional irrelevant objects are placed on the tabletop or in the surrounding scene, such as extra fruit items, plush bears, and other non-target objects. In the tablecloth setting, we change the tablecloth color and texture to alter the workspace appearance. In the lighting setting, we turn on dynamic colored lights to create non-stationary illumination changes and color shifts. These perturbations are designed to test whether the policy can focus on the task-relevant object, contact region, and intended scene change rather than relying on background appearance.

\subsubsection{Real-Robot Evaluation Protocol}
\label{app:real_eval_protocol}
\vspace{-0.8em}
Evaluation contains one Easy track and three Hard tracks. Easy uses 10 trials per task with in-distribution object poses and lighting. Each Hard track uses 5 trials per task and isolates one visual shift: distractor objects, lighting generalization, or tablecloth generalization. All methods are evaluated from matched initial states with the same language instruction sequence and the same manual reset procedure. On the Dobot robot, for binary tasks, a trial receives 1 for success and 0 otherwise. \texttt{PushButtons} receives $0$, $0.5$, or $1$ according to the number of correctly pressed targets. \texttt{CollectFruits} and \texttt{StackBowls} receive $0$, $0.3$, $0.6$, or $1$ according to task-specific completion stages. The Hard score is the sum over the three hard tracks, normalized as an average success rate. On the Franka robot, during evaluation, each trial is scored by stage, with a value between 0 and 1 assigned according to the successfully completed stages.

\section{Experimental Details}
\label{app:experimental_details}

\subsection{Implementation Details}
\label{app:implementation_details}
We train our models using 8 NVIDIA A100 80GB GPUs and 4 NVIDIA RTX PRO 6000 GPUs. We instantiate the VLM backbones with Florence-2~\citep{xiao2024florence}, PaliGemma~\citep{beyer2024paligemma}, and Qwen3.5~\citep{qwen35blog}, following X-VLA~\citep{zheng2025xvla}, $\pi_{0.5}$~\citep{black2025pi05}, and InternVLA-A1.5~\citep{ma2026internvla15}, respectively. The corresponding VLM sizes are 0.9B, 3B, and 2B parameters. We use AdamW with $\beta_1=0.9$, $\beta_2=0.95$, $\epsilon=10^{-8}$, and weight decay $0.01$. We clip the gradient norm to $1.0$ and train with BF16 mixed precision.

\subsection{Baselines}
\label{app:baselines}

\begingroup
\setlength{\parskip}{0pt}

\subsubsection{Vision-Language-Action Models}

\noindent\textbf{X-VLA}~\citep{zheng2025xvla} is a scalable flow-matching VLA built from standard Transformer encoders. It assigns each data source an embodiment-specific learned soft prompt, allowing a shared policy to exploit heterogeneous cross-robot data with only a small number of added parameters.

\noindent\textbf{$\boldsymbol{\pi}_{0.5}$}~\citep{black2025pi05} extends $\pi_0$ toward open-world manipulation through heterogeneous co-training. It combines multi-robot actions with web data, object detections, high-level semantic prediction, and hybrid multimodal examples to improve long-horizon generalization in unseen environments.

\noindent\textbf{LingBot-VLA}~\citep{wu2026linbotvla} is a dual-arm foundation policy trained on roughly 20K hours of real-world data from nine robot configurations. Its emphasis is practical cross-platform generalization, efficient post-training, and a high-throughput open training and deployment stack.


\noindent\textbf{InternVLA-A1.5}~\citep{ma2026internvla15} preserves VLM semantics through continued VQA and subtask-prediction training while attaching a lightweight expert for continuous actions. Learnable foresight tokens distill task-relevant future information from a frozen video generator, whose prediction branch is removed at deployment.

\subsubsection{World Action Models}


\noindent\textbf{Fast-WAM}~\citep{yuan2026fastwam} retains video-prediction co-training but removes explicit future imagination at test time. It isolates the representational benefit of world-model supervision from the cost of iterative video generation, enabling real-time action prediction without discarding future-aware training.




\noindent\textbf{VLA-JEPA}~\citep{sun2026vlajepa} pretrains an action policy through leakage-free latent future prediction: the student observes only the current frame, while a target encoder converts future frames into supervision. Predicting action-relevant state transitions in latent space reduces sensitivity to camera motion, background changes, and other pixel-level nuisance variation.


\noindent\textbf{Cosmos Policy}~\citep{kim2026cosmos} adapts a pretrained video diffusion model to control in a single post-training stage without adding a separate action architecture. Actions, future observations, and values are represented as latent frames, enabling the same model to generate trajectories and perform value-guided test-time planning.



\noindent\textbf{Motus}~\citep{bi2026motus} unifies understanding, video generation, and action experts in a Mixture-of-Transformers architecture with a shared diffusion scheduler. Optical flow supplies embodiment-agnostic latent-action supervision, allowing the model to switch among world modeling, inverse dynamics, VLA control, and joint video--action prediction.




\endgroup

\section{Additional Experimental Results and Analysis}
\label{app:additional_experimental_results}

\subsection{Detailed Real-Robot Results}

For the Franka robot, we evaluate the Easy track. The per-task success rates for \method{}, $\pi_{0.5}$~\citep{black2025pi05}, and VLA-JEPA~\citep{sun2026vlajepa} are reported in Tab.~\ref{tab:app_franka_easy_per_task}. Tab.\ref{tab:dobot_real} details the success rate of different models for each task and each track on the Dobot robot.

\begin{table*}[h]
\centering
\scriptsize
\setlength{\tabcolsep}{4pt}
\caption{Per-task success rates (\%) on the Franka real-robot evaluation.}
\resizebox{\textwidth}{!}{%
\begin{tabular}{lcccccc}
\toprule
Method & Collect Fruits & Press Buttons & Open Lip & Fold Towel &
Put Item In Drawer & Average \\
\midrule

$\pi_{0.5}$~\citep{black2025pi05}
& 50.0 & 80.0 & 50.0 & 30.0 & 25.0 & 47.0 \\

VLA-JEPA~\citep{sun2026vlajepa}
& \textbf{75.0} & 90.0 & 30.0 & 25.0 & 25.0 & 49.0 \\

\method{} (Ours)
& \textbf{75.0} & \textbf{100.0} & \textbf{60.0} & \textbf{80.0}
& \textbf{50.0} & \textbf{73.0} \\

\bottomrule
\end{tabular}%
}

\label{tab:app_franka_easy_per_task}
\end{table*}

\begin{table*}[t]
\centering
\scriptsize
\setlength{\tabcolsep}{2.5pt}
\renewcommand{\arraystretch}{1.05}
\caption{Dobot real-robot evaluation results (\%). Easy uses 10 trials per task; Hard contains three tracks with 5 trials each: distractors, lighting generalization, and tablecloth generalization. Each PushButtons test is scored as 0, 0.5, or 1; each CollectFruits and StackBowls test is scored as 0, 0.3, 0.6, or 1; all other tasks are scored as 0 or 1. Hard aggregates the results over the three hard tracks.}
\resizebox{\textwidth}{!}{%
\begin{tabular}{lccccccccccccccc}
\toprule
\multirow{2}{*}{\textbf{Method}} &
\multicolumn{5}{c}{\textbf{PickGrape}} &
\multicolumn{5}{c}{\textbf{StackBowls}} &
\multicolumn{5}{c}{\textbf{PushButtons}} \\
\cmidrule(lr){2-6}
\cmidrule(lr){7-11}
\cmidrule(lr){12-16}
& Easy & Dist. & Light & T-Cloth & Hard
& Easy & Dist. & Light & T-Cloth & Hard
& Easy & Dist. & Light & T-Cloth & Hard \\
\midrule

OpenVLA~\citep{kim2024openvla}
& 30.0 & 20.0 & 40.0 & 20.0 & 26.7
& 12.0 & 18.0 & 6.0 & 12.0 & 12.0
& \underline{80.0} & 70.0 & 40.0
& \textbf{80.0} & 63.3 \\

X-VLA~\citep{zheng2025xvla}
& \underline{80.0} & \underline{60.0} & \underline{80.0}
& \underline{60.0} & 66.7
& \underline{34.0} & \underline{24.0} & \underline{46.0}
& \underline{32.0} & \underline{34.0}
& \underline{80.0} & 70.0 & 40.0
& \textbf{80.0} & 63.3 \\

$\pi_{0.5}$~\citep{black2025pi05}
& \underline{80.0} & \underline{60.0} & \underline{80.0}
& \textbf{100.0} & \underline{80.0}
& \textbf{40.0} & \textbf{30.0} & \textbf{58.0}
& \textbf{42.0} & \textbf{43.3}
& \textbf{90.0} & \underline{80.0} & \underline{70.0}
& 60.0 & \underline{70.0} \\

\textbf{\model{} (Ours)}
& \textbf{90.0} & \textbf{80.0} & \textbf{100.0}
& \textbf{100.0} & \textbf{93.3}
& 30.0 & 18.0 & 40.0 & \textbf{42.0} & 33.3
& \textbf{90.0} & \textbf{90.0} & \textbf{90.0}
& \underline{70.0} & \textbf{83.3} \\

\midrule
\multirow{2}{*}{\textbf{Method}} &
\multicolumn{5}{c}{\textbf{CollectFruits}} &
\multicolumn{5}{c}{\textbf{PutItemInDrawer}} &
\multicolumn{5}{c}{\textbf{Avg. SR}} \\
\cmidrule(lr){2-6}
\cmidrule(lr){7-11}
\cmidrule(lr){12-16}
& Easy & Dist. & Light & T-Cloth & Hard
& Easy & Dist. & Light & T-Cloth & Hard
& Easy & Dist. & Light & T-Cloth & Hard \\
\midrule

OpenVLA~\citep{kim2024openvla}
& 20.0 & 12.0 & 18.0 & 12.0 & 14.0
& 10.0 & \underline{20.0} & 20.0 & 20.0 & 20.0
& 30.4 & 28.0 & 24.8 & 28.8 & 27.2 \\

X-VLA~\citep{zheng2025xvla}
& \underline{84.0} & \underline{72.0} & \textbf{100.0}
& \underline{66.0} & \underline{79.3}
& \textbf{70.0} & \textbf{40.0} & \textbf{60.0}
& \underline{40.0} & \textbf{46.7}
& 69.6 & 53.2 & 65.2
& 55.6 & 58.0 \\

$\pi_{0.5}$~\citep{black2025pi05}
& 80.0 & 66.0 & \underline{80.0}
& \underline{66.0} & 70.7
& \underline{60.0} & \textbf{40.0}
& \textbf{60.0} & \underline{40.0} & \textbf{46.7}
& \underline{70.0} & \underline{55.2} & \underline{69.6}
& \underline{61.6} & \underline{62.1} \\

\textbf{\model{} (Ours)}
& \textbf{88.0} & \textbf{86.0} & \textbf{100.0}
& \textbf{80.0} & \textbf{88.7}
& \textbf{70.0} & \textbf{40.0} & \underline{40.0}
& \textbf{60.0} & \textbf{46.7}
& \textbf{73.6} & \textbf{62.8} & \textbf{74.0}
& \textbf{70.4} & \textbf{69.1} \\

\bottomrule
\end{tabular}%
}

\label{tab:dobot_real}
\end{table*}

\subsection{Ablation Study}
\label{app:Future-Prediction-Transfer-Analysis}

\begin{figure}[!t]
    \centering
    \includegraphics[width=1\linewidth]{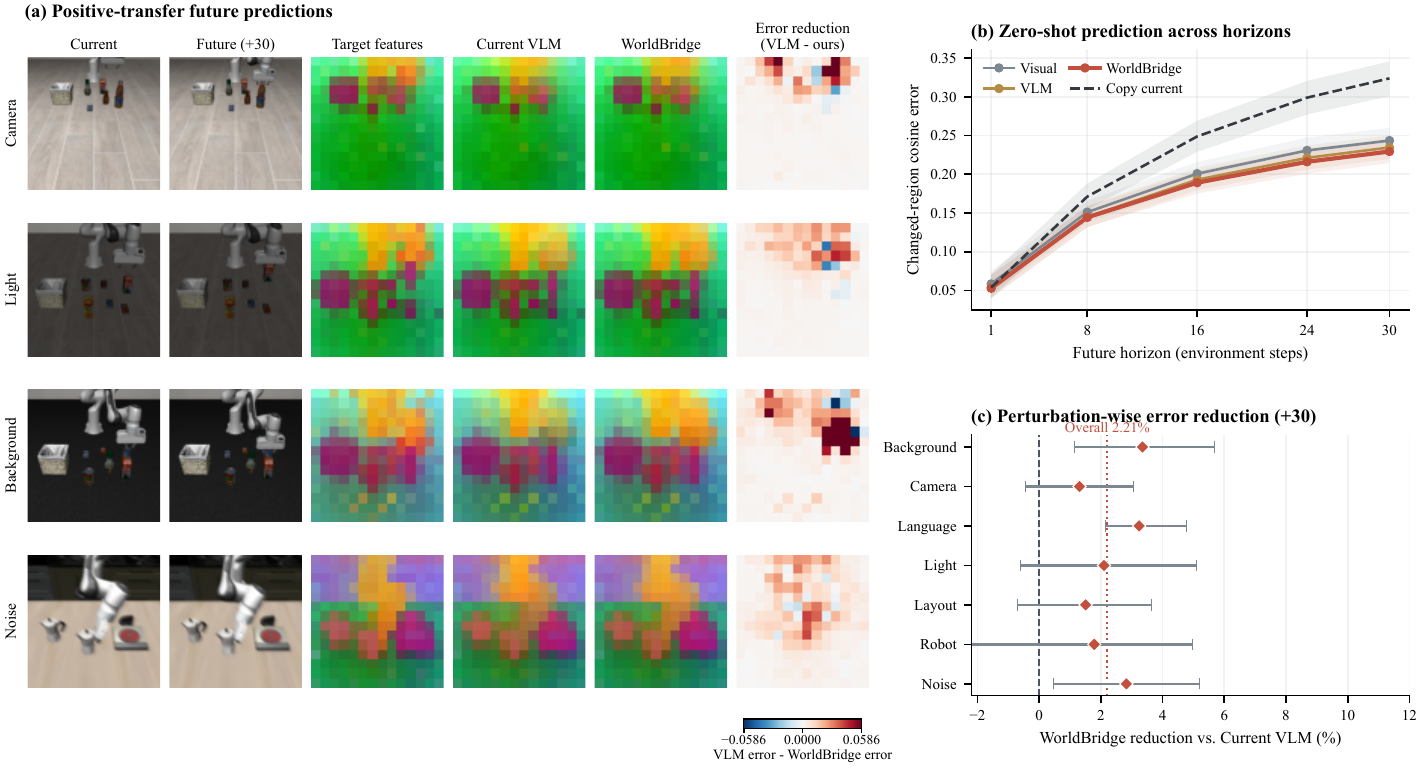}
    \vspace{-18pt}
    \caption{\textbf{Future-feature prediction under LIBERO-Plus perturbations.}
(a) Positive-transfer examples for Camera, Light, Background, and Sensor Noise. Red regions indicate lower future-feature prediction error for WorldBridge than for Current VLM.
(b) Changed-region cosine error across future horizons.
(c) WorldBridge error reduction relative to Current VLM at horizon $+30$, with task/episode-clustered 95\% confidence intervals. }
    \label{fig:future_prediction_main_combined}
\end{figure}

\paragraph{Future-Prediction Transfer Analysis} We investigate whether WorldBridge provides predictive information beyond the current VLM representation. As illustrated in Fig.~\ref{fig:future_prediction_main_combined}(a), WorldBridge more closely recovers the target future features, with improvements concentrated around regions undergoing future robot or object changes. These examples provide a spatial interpretation of the learned future priors, while the following quantitative results use the complete evaluation cohort. Equal-capacity diagnostic heads are trained only on original LIBERO and evaluated without adaptation on LIBERO-Plus. Figure~\ref{fig:future_prediction_main_combined}(b) shows that WorldBridge maintains lower changed-region prediction error than Current VLM across future horizons. At horizon $+30$, the error decreases from $0.2347$ to $0.2295$, corresponding to a relative reduction of $2.21\%$. Figure~\ref{fig:future_prediction_main_combined}(c) further reports positive point estimates across all seven perturbation families. The confidence intervals are entirely positive for Background, Language, and Sensor Noise. Overall, these results indicate that WorldBridge contributes transferable future-state information beyond the current VLM context.

\section{LIBERO-Dynamic Benchmark}
\label{app:libero_dynamic}

\subsection{Motivation and Benchmark Definition}
\label{app:libero_dynamic_definition}

Existing LIBERO evaluations mainly alter the initial environment state or apply
a fixed visual perturbation throughout an episode. Such protocols measure
robustness to static distribution shifts, but do not fully evaluate whether a
policy can maintain temporally consistent perception and control when the
observation viewpoint changes during task execution.

\begin{figure}
    \centering
    \includegraphics[width=1\linewidth]{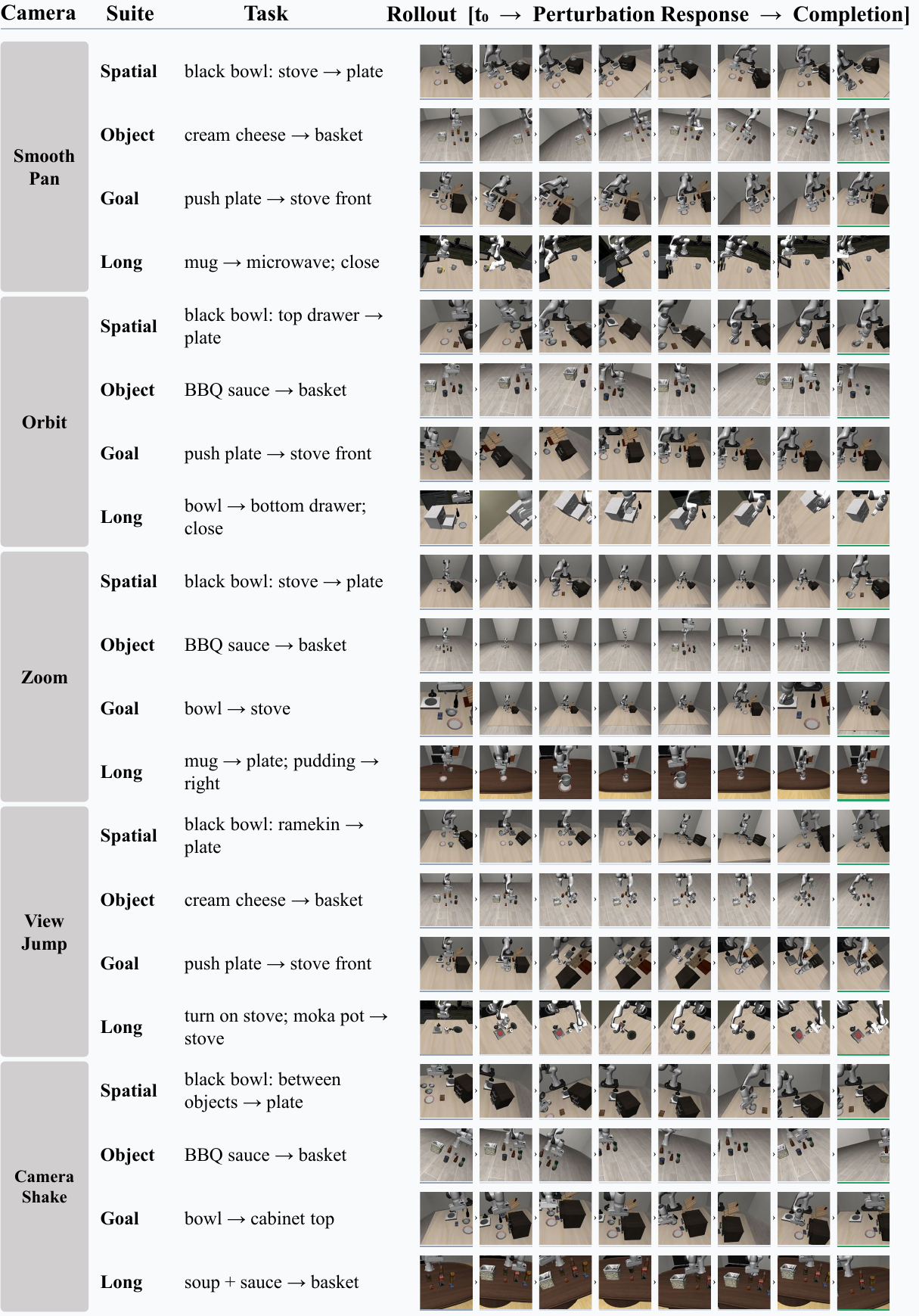}
    \caption{\textbf{Qualitative examples from LIBERO-Dynamic.}
Representative rollout sequences under five dynamic camera variants:
Smooth Pan, Orbit, Zoom, View Jump, and Camera Shake. For each variant, we
show tasks from the Spatial, Object, Goal, and Long suites. Frames are ordered
from left to right, beginning with the initial observation at $t_0$, followed
by the time-varying camera perturbation and the policy response, and ending
with task completion.}
    \label{fig:libero_dynamic_vis}
\end{figure}

We introduce \textbf{LIBERO-Dynamic}, as shown in the Fig\ref{fig:libero_dynamic_vis}, a zero-shot evaluation benchmark for
temporally varying camera perturbations. Unlike static camera benchmarks,
LIBERO-Dynamic moves the external agent-view camera as the robot executes the
task. The manipulation task, object dynamics, robot state, language
instruction, and success condition remain unchanged. Therefore, performance
degradation can be attributed primarily to the policy's robustness to dynamic
viewpoint changes.

Let $\mathcal{T}$ denote a LIBERO task, $s_i$ its $i$-th initial environment
state, and $H_{\mathcal{T}}$ its maximum episode horizon. For camera variant
$v$, LIBERO-Dynamic assigns a deterministic camera trajectory
\begin{equation}
    \Gamma_{\mathcal{T},i,v}
    =
    \left\{
    \mathbf{C}_{\mathcal{T},i,v}(t)
    \right\}_{t=0}^{H_{\mathcal{T}}-1},
\end{equation}
where $\mathbf{C}(t)=(\mathbf{p}(t),\mathbf{R}(t))$ contains the camera position
and orientation at simulation step $t$. We parameterize time using
\begin{equation}
    u_t = \frac{t}{H_{\mathcal{T}}-1},
\end{equation}
so that the camera motion spans the complete task horizon and remains
comparable across tasks with different maximum lengths.

Each tuple $(\mathcal{T},s_i,v)$ maps to one unique but deterministic
trajectory. The mapping is fixed in a released evaluation manifest rather than
sampled online. Consequently, different models observe exactly the same
initial state and camera motion, enabling paired and reproducible comparison.

\paragraph{Difference from static camera evaluation.}
LIBERO-Dynamic evaluates three capabilities that are weakly measured by a
fixed viewpoint: temporal visual consistency, recovery from changing spatial
projections, and action prediction under continuously changing
camera--workspace geometry. It is an evaluation-only benchmark: models are
trained on the standard LIBERO data and evaluated on LIBERO-Dynamic without
dynamic-camera training data or benchmark-specific fine-tuning.

\subsection{Dynamic Camera Variants}
\label{app:libero_dynamic_variants}

LIBERO-Dynamic contains five complementary camera-motion variants. They cover
continuous, discontinuous, low-frequency, and high-frequency viewpoint
changes.

\begin{table}[t]
    \centering
    \small
    \setlength{\tabcolsep}{4.5pt}
    \caption{Dynamic camera variants in LIBERO-Dynamic.}
    \begin{tabular}{p{0.20\linewidth}p{0.72\linewidth}}
        \toprule
        \textbf{Variant} & \textbf{Description} \\
        \midrule
        Smooth Pan &
        Applies smooth low-frequency changes in azimuth and elevation. A
        subset of trajectories follows an arc around the workspace while
        continuously looking toward its center. \\
        Orbit &
        Moves the camera along a continuous orbit around the workspace center.
        The look-at direction is updated at every simulation step to preserve
        the manipulation region in view. \\
        Zoom &
        Changes the camera--workspace distance along the viewing direction,
        producing temporally varying object scale and perspective while
        retaining the task-relevant workspace. \\
        View Jump &
        Switches between several separated, valid viewpoints during task
        execution. Some jumps are sampled from center-looking orbital poses.
        This variant evaluates recovery from abrupt changes in visual
        correspondence. \\
        Camera Shake &
        Adds band-limited translational and rotational perturbations at a
        higher temporal frequency, simulating unstable camera mounting or
        external vibration. \\
        \bottomrule
    \end{tabular}
    
    \label{tab:libero_dynamic_variants}
\end{table}

A camera trajectory is accepted only when it satisfies workspace-visibility
constraints. In particular, the camera height, viewing direction, distance,
and angular range are bounded, and center-looking trajectories keep the
workspace center inside a valid image region. Invalid poses that point toward
the floor, ceiling, or outside the workspace are rejected or projected back
into the valid camera region. These constraints prevent degenerate samples in
which the task becomes unobservable rather than perceptually challenging.

\subsection{Difficulty Levels}
\label{app:libero_dynamic_difficulty}

Each variant is divided into five ordered difficulty levels. Difficulty is
increased by jointly scaling the translational displacement, angular
excursion, motion speed, active duration, jump distance or frequency, and
shake magnitude, depending on the variant. The levels are defined such that
\begin{equation}
    A_1 < A_2 < A_3 < A_4 < A_5,
\end{equation}
where $A_d$ denotes the effective perturbation magnitude at difficulty level
$d$. Exact trajectory parameters are stored in the frozen benchmark manifest.


\subsection{Evaluation Protocol}
\label{app:libero_dynamic_protocol}

We evaluate all 40 tasks from four standard LIBERO suites: LIBERO-Spatial, LIBERO-Object, LIBERO-Goal, and LIBERO-10. Each suite contains ten tasks. Every task is evaluated from 50 initial states under each of the five camera variants. Thus, one variant contains 2000 rollouts, and the complete benchmark contains 10000 rollouts. We report the success rate for each camera variant and the overall average.

\section{Additional Visualizations}
\label{app:additional_visualizations}

\begin{figure}[h]
    \includegraphics[width=1\linewidth]{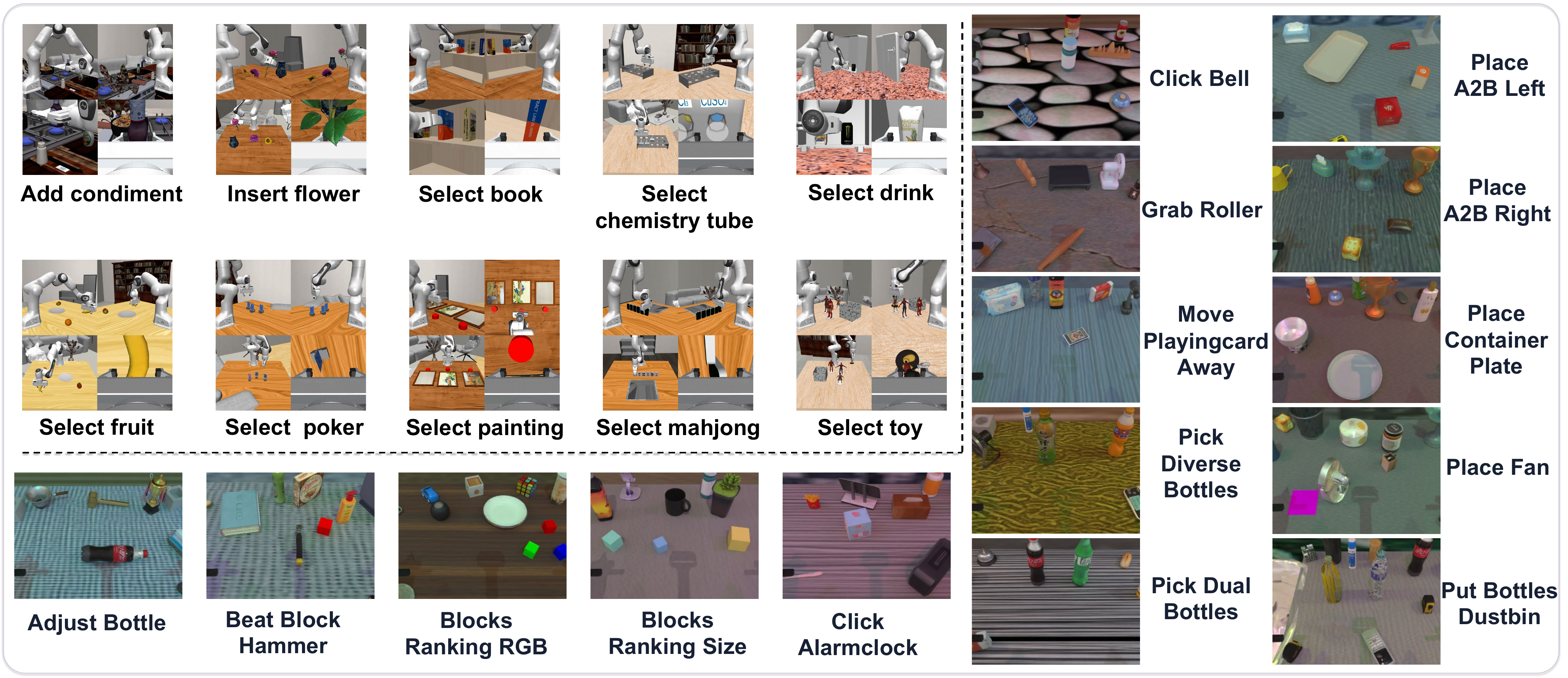}
    \vspace{-15pt}
    \caption{Task and scene visualizations for the simulation benchmarks on VLABench and RoboTwin~2.0.}
    \label{fig:app_sim_task_gallery}
\end{figure}

\begin{figure}
    \includegraphics[width=1\linewidth]{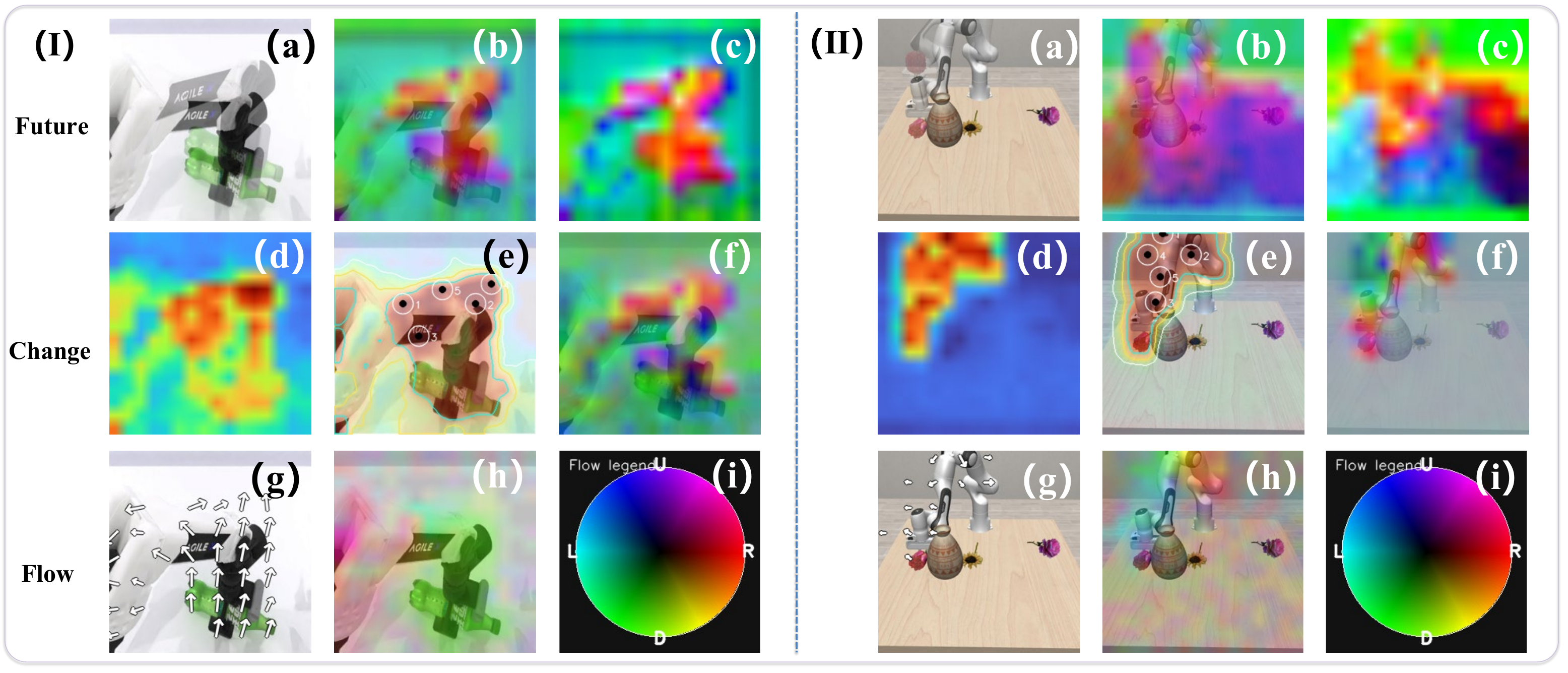}
    \vspace{-15pt}
    \caption{Qualitative world-prior visualizations on RoboTwin~2.0 (\uppercase\expandafter{\romannumeral 1}) and VLABench (\uppercase\expandafter{\romannumeral 2}). We show the factorized predictive signals used by \method{}: future representations summarize the intended outcome, change maps identify action-relevant regions expected to change, and motion-flow maps indicate the local direction of scene evolution. (a) shows the future RGB target overlaid on the current observation. 
(b) overlays the predicted future latent representation on the scene, while (c) visualizes the future latent map. 
(d) presents the change map, and (e) overlays the enhanced change response on the observation; the marked points indicate high-confidence regions where action-relevant state changes are expected. 
(f) highlights feature-level differences between the current and future states. 
(g) visualizes the dense flow field on the scene, and (h) illustrates the temporal motion direction implied by the flow.
(i) gives the color-wheel legend, where hue denotes motion direction. }
    \label{fig:app_sim_world_prior_vis}
\end{figure}

\begin{figure}
    \includegraphics[width=1\linewidth]{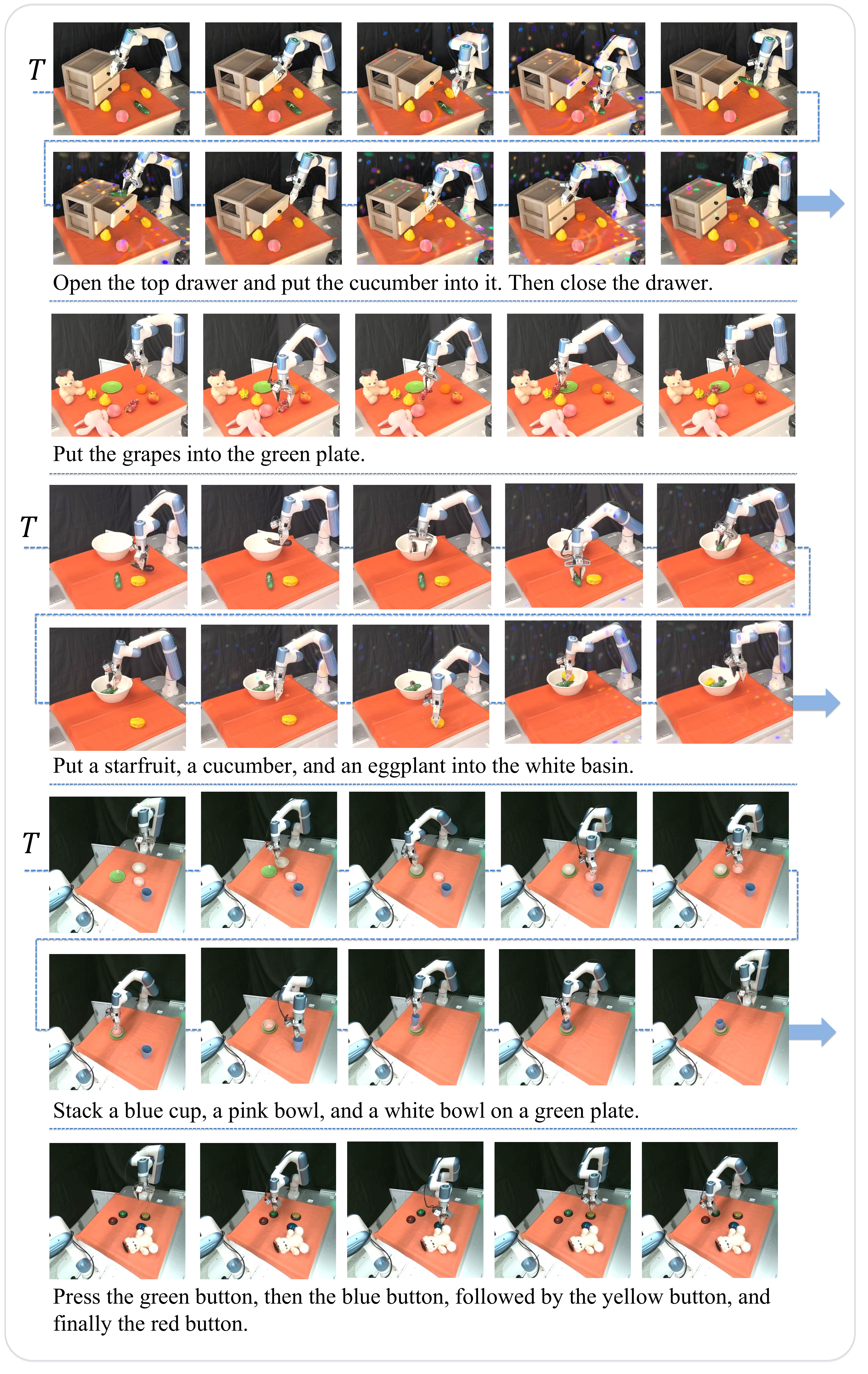}
    \vspace{-15pt}
    \caption{Dobot real-robot execution sequences. The figure shows temporal rollouts for the five real-robot tasks, covering object selection, placement, pushing, stacking, and drawer manipulation.}
    \label{fig:app_dobot_task_process}
\end{figure}

We provide additional qualitative visualizations to contextualize the benchmark tasks, the learned world-prior representations, and the real-robot execution behavior. The benchmark-scene overview in Fig.~\ref{fig:app_sim_task_gallery} visualizes the ten VLABench tasks and the 15 RoboTwin~2.0 tasks. The world-prior visualization in Fig.~\ref{fig:app_sim_world_prior_vis} illustrates how \method{} decomposes predicted scene evolution into future, change, and motion priors on simulation benchmarks. Finally, the Dobot rollout sequence in Fig.~\ref{fig:app_dobot_task_process} presents representative executions for the five real-robot tasks, showing the temporal progression from initial observation to task completion.

\end{document}